%% file: main.tex
\documentclass[10pt,journal,compsoc]{IEEEtran}

\ifCLASSOPTIONcompsoc
  \usepackage[nocompress]{cite}
\else
  \usepackage{cite}
\fi

\usepackage{graphicx}

\newcommand{\etal}{\textit{et al.}}
\newcommand{\eg}{\textit{e.g.,}}
\newcommand{\specialcell}[2][c]{\begin{tabular}[#1]{@{}c@{}}#2\end{tabular}}

\usepackage{xcolor}

\usepackage{amsmath}
\usepackage{amssymb}

\usepackage{threeparttable}
\usepackage{booktabs}
\usepackage{csquotes}
\usepackage[caption=false,font=footnotesize]{subfig}
\usepackage{float}
\usepackage{tabularx}
\usepackage{booktabs}
\usepackage{multirow}
\usepackage{caption}
\usepackage{footmisc}
\usepackage{xspace}
\usepackage[numbers,sort&compress]{natbib}

\PassOptionsToPackage{hyphens}{url}\usepackage[pagebackref=true,breaklinks=true,letterpaper=true,bookmarks=false]{hyperref}

\makeatletter
\DeclareRobustCommand\onedot{\futurelet\@let@token\@onedot}
\def\@onedot{\ifx\@let@token.\else.\null\fi\xspace}

\def\eg{\emph{e.g}\onedot,~} 
\def\ie{\emph{i.e}\onedot,~} 
 
\def\etc{\emph{etc}\onedot} 
 
\def\etal{\emph{et al}\onedot}
\makeatother

\begin{document}

\title{Biometrics: Trust, but Verify}

\author{Anil~K.~Jain,~\IEEEmembership{Life Fellow,~IEEE,}
        Debayan~Deb,~\IEEEmembership{Student Member,~IEEE,}
        and~Joshua~J.~Engelsma,~\IEEEmembership{Student Member,~IEEE}
}

\markboth{Journal of \LaTeX\ Class Files,~Vol.~14, No.~8, August~2015}%
{Jain \MakeLowercase{\textit{et al.}}: Biometrics for Trust and Trust for Biometrics}

\IEEEtitleabstractindextext{%
\begin{abstract}

\input{ch_abstract}

\end{abstract}

}


\twocolumn[{%
\renewcommand\twocolumn[1][]{#1}%
\begin{@twocolumnfalse}
\maketitle
\IEEEdisplaynontitleabstractindextext
   \IEEEpeerreviewmaketitle
  \end{@twocolumnfalse}

\begin{center}
    \centering
    \captionsetup{font=footnotesize}
    \captionsetup[subfloat]{justification=centering}
    \begin{minipage}{\textwidth}
    \begin{figure}[H]
    \centering
    \vspace{-4em}
    \subfloat[Surveillance~\cite{surveillance}]{
        \centering
        \includegraphics[width=0.19\textwidth]{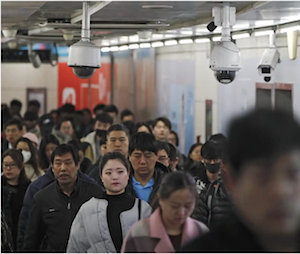}
    }
     \subfloat[e-Commerce~\cite{ecommerce}]{
        \centering
        \includegraphics[width=0.19\textwidth]{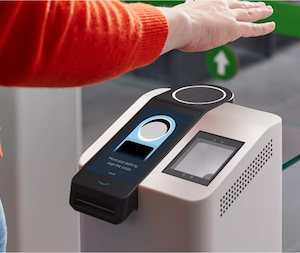}
    }
     \subfloat[Airport Security~\cite{boarding}]{
        \centering
        \includegraphics[width=0.19\textwidth]{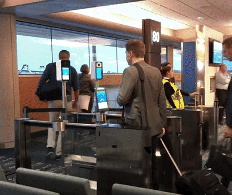}
    }
     \subfloat[Match-on-Card~\cite{matchoncard}]{
        \centering
        \includegraphics[width=0.19\textwidth]{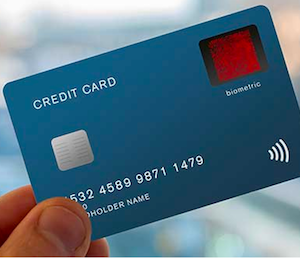}
    }
     \subfloat[Biometric ATM~\cite{atm}]{
        \centering
        \includegraphics[width=0.19\textwidth]{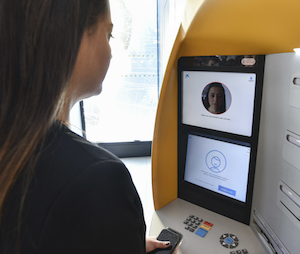}
    }\\
    \subfloat[Smartphone Access~\cite{smartphone}]{
        \centering
        \includegraphics[width=0.19\textwidth]{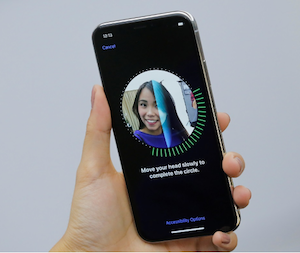}
    }
     \subfloat[Border Control~\cite{bordercontrol}]{
        \centering
        \includegraphics[width=0.19\textwidth]{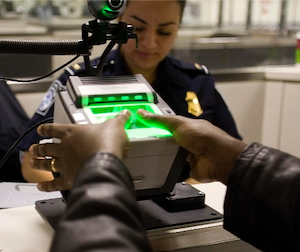}
    }
     \subfloat[Social Welfare Benefits~\cite{socialbenefits}]{
        \centering
        \includegraphics[width=0.19\textwidth]{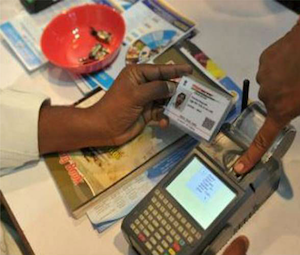}
    }
     \subfloat[Time and Attendance~\cite{coalminers}]{
        \centering
        \includegraphics[width=0.19\textwidth]{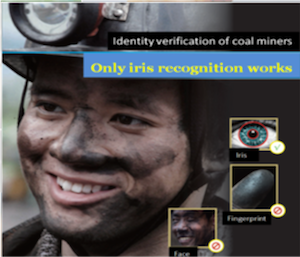}
    }
     \subfloat[Vehicular Biometrics~\cite{vehicular}]{
        \centering
        \includegraphics[width=0.19\textwidth]{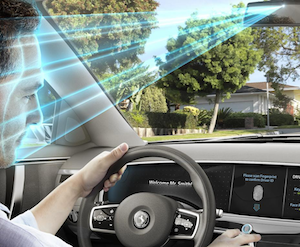}
    }
    
    \captionof{figure}{Examples where biometrics are introduced for trust. For instance, Amazon employs~\emph{Amazon One}, a biometric recognition system for e-commerce, that lets shoppers pay for their groceries by authenticating them via their palmprints~\cite{amazon_one_news}. US-VISIT authenticates international travelers to the United States via their fingerprints~\cite{us_visit}. ``Touchless" authentication via face recognition is being increasingly employed for entry, exit, and flight boarding~\cite{boarding} for airport security.}
    \label{fig:biometric_applications}
    \end{figure}
    \vspace{1em}
    \end{minipage}
\end{center}
}]

{
  \renewcommand{\thefootnote}%
    {\fnsymbol{footnote}}
\footnotetext{\textit{A.~K.~Jain, D.~Deb and J.~J.~Engelsma are with the Department of Computer Science and Engineering, Michigan State University, East Lansing, MI, 48824.}\protect\\
\textit{E-mail: \{jain, debdebay, engelsm7\}@msu.edu}
}
}

\input{ch_intro}
\input{ch_performance}
\input{ch_security}
\input{ch_interpret}
\input{ch_bias}
\input{ch_privacy}
\input{ch_conclusion}

\ifCLASSOPTIONcaptionsoff
  \newpage
\fi

\bibliographystyle{ieeetr}
\footnotesize
\bibliography{egbib}

\begin{IEEEbiography}[{\includegraphics[width=1in,height=1.25in,clip,keepaspectratio]{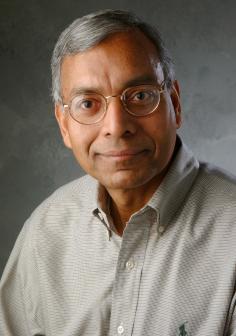}}]{Anil K. Jain}
is a University distinguished professor in the Department of Computer Science and Engineering at Michigan State University. His research interests include pattern recognition and biometric authentication. He served as the editor-in-chief of the IEEE Transactions on Pattern Analysis and Machine Intelligence and was a member of the United States Defense Science Board. He has received Fulbright, Guggenheim, Alexander von Humboldt, and IAPR King Sun Fu awards. He is a member of the National Academy of Engineering, and The World Academy of Sciences, and foreign fellow of the Indian National Academy of Engineering and Chinese Academy of Sciences.
\end{IEEEbiography}

\begin{IEEEbiography}[{\includegraphics[width=1in,height=1.25in,clip,keepaspectratio]{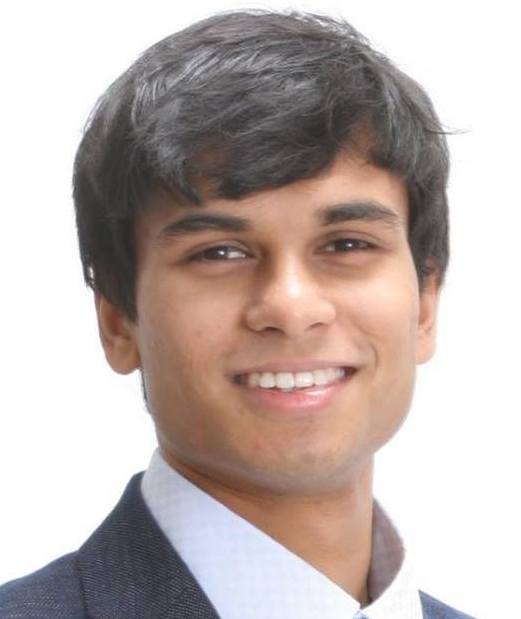}}]{Debayan Deb}
received his B.S. degree in  computer  science  from  Michigan State University,  East Lansing,  Michigan,  in  2016.  He  is currently working towards a PhD degree in the Department of Computer Science and Engineering at Michigan State University, East Lansing, Michigan. His research interests include pattern recognition, computer vision, and machine learning with applications in biometrics.
\end{IEEEbiography}


\begin{IEEEbiography}[{\includegraphics[width=1in,height=1.25in,clip,keepaspectratio]{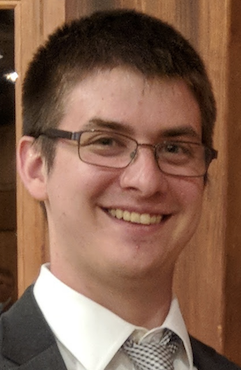}}]{Joshua J. Engelsma}
graduated magna cum laude with a B.S. degree in computer science
from Grand Valley State University, Allendale,
Michigan, in 2016. He is currently working towards a PhD degree in the Department of Computer Science and Engineering at Michigan
State University. His research interests include
pattern recognition, computer vision, and image
processing with applications in biometrics. He
won the best paper award at the 2019 IEEE International Conference on Biometrics (ICB), and the 2020 Michigan State University College of Engineering Fitch Beach
Award.
\end{IEEEbiography}

\end{document}

%% file: ch_abstract.tex
Over the past two decades, biometric recognition has exploded into a plethora of different applications around the globe. This proliferation can be attributed to the high levels of authentication accuracy and user convenience that biometric recognition systems afford end-users. However, in-spite of the success of biometric recognition systems, there are a number of outstanding problems and concerns pertaining to the various sub-modules of biometric recognition systems that create an element of mistrust in their use - both by the scientific community and also the public at large. Some of these problems include: i) questions related to system recognition performance, ii) security (spoof attacks, adversarial attacks, template reconstruction attacks and demographic information leakage), iii) uncertainty over the bias and fairness of the systems to all users, iv) explainability of the seemingly black-box decisions made by most recognition systems, and v) concerns over data centralization and user privacy. In this paper, we provide an overview of each of the aforementioned open-ended challenges. We survey work that has been conducted to address each of these concerns and highlight the issues requiring further attention. Finally, we provide insights into how the biometric community can address core biometric recognition systems design issues to better instill trust, fairness, and security for all.

%% file: ch_intro.tex
\IEEEraisesectionheading{\section{Introduction}\label{sec:introduction}}

\IEEEPARstart{T}{he} Digital Age we live in has accelerated a proliferation of sensitive and personal data needing absolute protection. For instance, most of us now carry access to our bank account, email, business dealings, private message history, personal videos and photos, and much more all within a few taps on the smartphones in our pockets. It goes without saying that such data needs to be  secured at all times. At the same time, users want the convenience of being able to access such data in a seamless and safe manner. It is therefore not surprising that virtually all smartphones now come equipped with a biometric authentication system (either face or fingerprint) for highly \textit{accurate} and \textit{convenient} unlocking of our phones. In addition, every day, a variety of organizations pose identity-related questions such as,~\emph{Should John be granted a visa?},~\emph{Does Alice already have a driver's license?}, and~\emph{Is Cathy the owner of the bank account?} Consequently, the use of biometric recognition systems has now pervaded into the lives of billions of human-beings all around the globe through a variety of applications (Figure~\ref{fig:biometric_applications}). 

Biometric recognition, or simply biometrics, refers to~\emph{automatic} person recognition based on an individual's physical or behavioral traits~\cite{jain2011introduction}. The term,~\emph{Biometrics}, is derived from the Greek words~\emph{bios} (life) and~\emph{metron} (measure). Hence, biometrics in the context of person recognition refers to recognition based on measurements of the body (\eg face, fingerprint and iris). The origin of modern day biometric recognition has its roots in the ``Habitual Criminals Act" passed by the British Parliament in 1869~\cite{habitual_criminals}. In particular, the Home Office Committee expressed the need for a reliable person recognition scheme for tracing repeat offenders~\cite{home_office_quote},
\begin{quote}
    \textit{``What is wanted is a means of classifying the records of habitual criminal, such that as soon as the particulars of the personality of any prisoner (whether description, measurements, marks, or photographs) are received, it may be possible to ascertain readily, and with certainty, whether his case is in the register, and if so, who he is."}~\cite{home_office_quote}
\end{quote}

\noindent In essence, biometrics relies on who you are or how you act as opposed to what you know (such as a password) or what you have (such as an ID card).

Prior to automated biometric recognition systems, reliable identification of fellow beings had been a long-standing problem in human society. In early civilizations, people lived in small, connected communities. However, as humanity became more mobile and populations increased, we needed to start relying on credentials for person recognition. Dating back all the way to ancient Rome, passwords had long been viewed as the ideal method of securing information and gaining access to exclusivity~\cite{ancient_rome}. While passwords may have served their purpose in ancient Rome, in this day and age, passwords, while still in common use, are rife with problems. For example, passwords are prone to social engineering hacks, where someone can access a user's password by gaining their trust~\cite{social_engineering}. Alternatively, a malicious individual can observe and log a victim's typed password characters on a keyboard~\cite{keylogging}. Finally, plain-text passwords may be hacked or leaked from an insecure database~\cite{msn}. Other knowledge-based authentication schemes such as PINs are also prone to such attacks~\cite{smudge_attacks}. To combat the limitations imposed by passwords, an alternative authentication scheme involves physical tokens, such as certificates, ID cards, passports and driver's licenses. Unfortunately, these tokens are also vulnerable to social engineering attacks and theft. Furthermore, in developing countries around the world, many economically disadvantaged individuals lack any type of identification documentation making it difficult for them to access government benefits, healthcare, and financial services. If an individual does possess an official ID document, it may be fraudulent or shared with others~\cite{fraud, fraud2, fraud3}. Finally, even if identification documentation can be adequately distributed to everyone in a society, it cannot be trusted. For example, Dhiren Barot, an Al-Qaeda fanatic, was issued with nine fake British passports~\cite{dhiren}. 

Not surprisingly, the problems associated with password or token based authentication and identification has led to society exploring a more accurate and reliable method of user authentication and identification management systems which society as a whole can \textit{trust}. The word \textit{``trust"} is defined in the Oxford dictionary as~\cite{stevenson2010oxford}:

\begin{quote}
\textbf{TRUST:}~\textit{``Firm belief in the \textbf{reliability}, \textbf{truth}, \textbf{ability}, or strength of someone or something."}
\end{quote}

\noindent Thus for biometric recognition to be used in lieu of conventional passwords or as an identity management system, they must be shown to be highly accurate (establishing the reliability and truth portion of the definition) and also robust, or reliable. In other words, biometric recognition systems must be demonstrated to be \textit{trustworthy}. Subsequently and finally, a \textit{firm belief} in this trustworthiness must be established with system users to gain their trust.

To date, significant progress has been made in solidifying the \textit{accuracy} component of a trustworthy biometric recognition system. In particular, while automated biometric recognition systems have now been around for quite some time\footnote{Trauring's landmark paper on automated fingerprint recognition~\cite{trauring1963automatic} appeared in 1963, while the first Automated Fingerprint Identification Systems (AFIS) became available only in mid 1980s~\cite{iafis}.}, recent advances in hardware (\eg an NVIDIA 3090 GeForce RTX performs at 35.58 TFLOPS\footnote{\url{https://www.nvidia.com/en-ph/geforce/graphics-cards/30-series/rtx-3090/}}) and computer vision algorithms (specifically deep learning~\cite{deep_learning_biometrics1, deep_learning_biometrics2, deep_learning_biometrics3}) have led to biometric recognition systems which now surpass human recognition performance~\cite{lu2015surpassing}. More specifically, NIST evaluations for fingerprint~\cite{nist_fpvte}, face~\cite{grother2018ongoing}, and iris~\cite{quinn2019irex} search algorithms boast accuracies of FNIR = $0.001$, $0.058$, and $0.0059$ @ FPIR = $0.001$, respectively (Table~\ref{table:benchmark_sota_id}).

\begin{table}[t]
\centering
\caption{State-of-the-art identification (search) accuracy for Fingerprint, Face, and Iris.}
\label{table:benchmark_sota_id}
\begin{threeparttable}
\resizebox{\linewidth}{!}{\begin{tabular}{cccc}
\toprule
Trait & Evaluation & Gallery Size & Iden. Error\tnote{1}\\
\midrule
Fingerprint & NIST FpVTE 2012 & 5M\tnote{2} & 0.001\\
\midrule
Face & Ongoing NIST FRVT & 12M & 0.058\\
\midrule
Iris & NIST IREX 10 & 500K & 0.006\\
\bottomrule
\end{tabular}}
\begin{tablenotes}
\item[1] FNIR @ FPIR = $0.001$.
\item[2] 10-print fusion performance.
\end{tablenotes}
\end{threeparttable}
\end{table}


Although the accuracy and convenience of biometric recognition systems has fueled their replacement of traditional password or token based methods (and more importantly, their widespread use in identity management systems), scientists must begin shifting their attention away from a purely recognition accuracy and convenience driven mindset to concerns voiced by policy makers and the general public about the \textit{reliability} of biometric recognition systems (first component of the definition of trust). Biometric systems are here to stay and their proliferation in our society will continue to grow. It is also given that biometric systems will make incorrect decisions, albeit small, and, like any security system, will be subjected to attacks by hackers. Therefore, the following concerns must be adequately addressed:

\begin{figure*}
    \centering
    \captionsetup{font=footnotesize}
    \includegraphics[width=\linewidth]{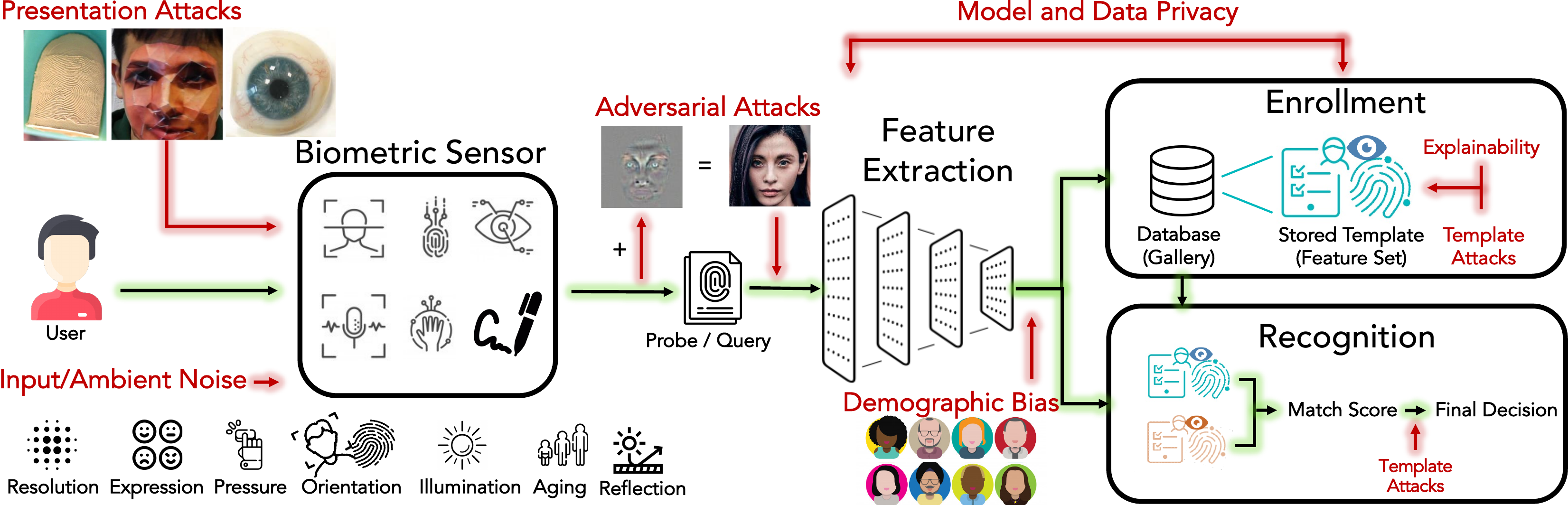}
    \caption{A typical biometric recognition pipeline (highlighted in green) consists of: (i) biometric sensor that generates a digital representation of a biometric trait, (ii) feature extractor that generates a compact and salient feature set, and (iii) matcher that outputs the final decision. We show the five major points that reduces trust in biometrics (highlighted in red): (i) robustness to adverse noises, (ii) biasness, (iii) security from biometric attacks, (iv) explainability, and (v) privacy.}
    \label{fig:pipeline}
\end{figure*}

\begin{enumerate}

\item \textbf{Performance: } Although biometric recognition system accuracy has matured, are there inputs and ambient noise that will still break the system? How will the recognition system perform over time? How will the system scale to millions or even billions of users? 

\item \textbf{Bias and Fairness: } Does the biometric recognition system work as well across all demographic groups? Does the system mis-classify members of one demographic group more than another (\eg age, gender, race, ethnicity and country of origin)? Why? What are the sources of bias in a biometric recognition system?

\item \textbf{Security: } Have biometric recognition systems solved the spoofing (presentation attack) vulnerability? Are biometric recognition systems robust to adversarial perturbations? Can users' templates stored in the system database be stolen or altered and used to reconstruct a biometric image or glean demographic information? How can we thwart these attack vectors?

\item \textbf{Explainability and Interpretability: } Why is the biometric recognition system making the decision it is making? What parts of the input image are being used to make a final decision? What features of the input image are most important in the decision? Will these features enable the model to operate accurately and consistently  over time and in different operating conditions?

\item \textbf{Privacy: } Even if we have a highly accurate and secure biometric system, how can we protect privacy of end users (and those who are in the training database)? Can we train on decentralized data, \eg~federated learning? Can we perform training or make inference directly on encrypted data? Can the model parameters also be encrypted?

\end{enumerate}

\noindent In other words, the trustworthiness of biometric recognition systems must be \textit{verified}~\cite{stevenson2010oxford}.

\begin{quote}
    \textbf{VERIFY:}~\textit{``The process of establishing the truth, accuracy, or validity of something."}
\end{quote}

\noindent While some work has begun to verify behaviors of biometric recognition systems via studying the aforementioned questions with scientific rigour, we argue that more work remains to be done. To that end, in this paper, we point out each of the major points of attack, question, or concern (Figure~\ref{fig:pipeline}) on the biometric recognition pipeline. Next, we systematically survey the literature to locate pertinent research aimed at addressing the aforementioned questions. We discuss remaining limitations left by the existing literature. Finally, we summarize recommended steps that can be taken and research that can be pursued (and also how it can be conducted rigorously, fairly, etc.) to build biometric recognition systems which are more trustworthy.

We note that this paper is unique in that it aggregates and examines the main components of a trustworthy biometric recognition system into one manuscript. Indeed many surveys~\cite{marasco2014survey, singh2020survey, jain2008biometric, jain2011introduction, jain_50, ross2019some, jain2015bridging, vakhshiteh2020adversarial, drozdowski2020demographic, boyd2020iris} have been written in great detail on each one of these topics individually, however we posit that there is benefit in extracting the key points from each of these areas and summarizing them in one place such that researchers can very quickly and easily assess the current state of trustworthy biometric recognition systems. Furthermore, many of the existing survey papers on these individual topics have become outdated. In short, this paper provides the \textit{latest} and most \textit{comprehensive} overview of the state of trustworthy biometric recognition systems.

%% file: ch_performance.tex
\begin{figure*}
    \centering
    \captionsetup{font=footnotesize}
    \subfloat[Fingerprint]{\includegraphics[height=1.3in]{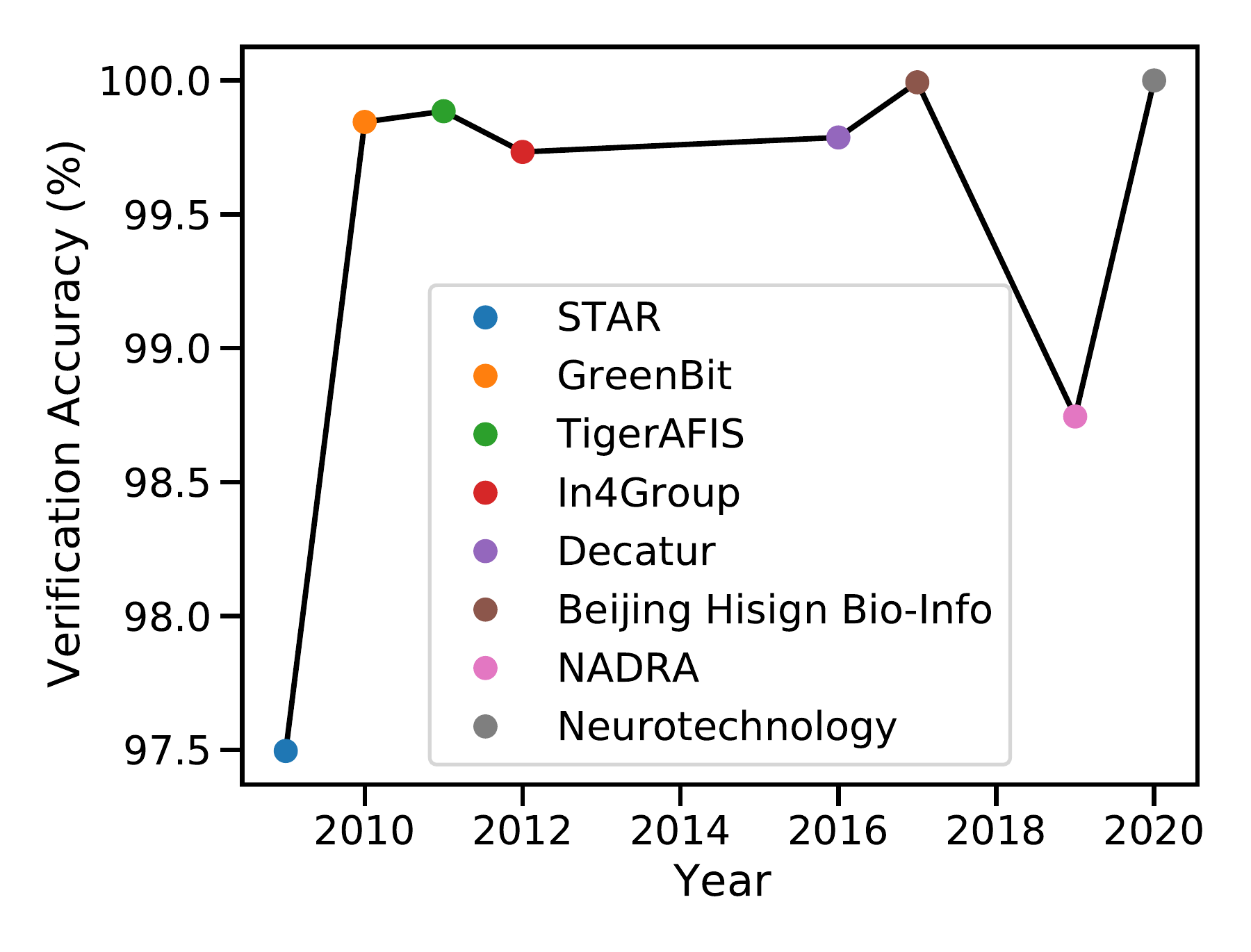}}\hfill
    \subfloat[Face]{\includegraphics[height=1.3in]{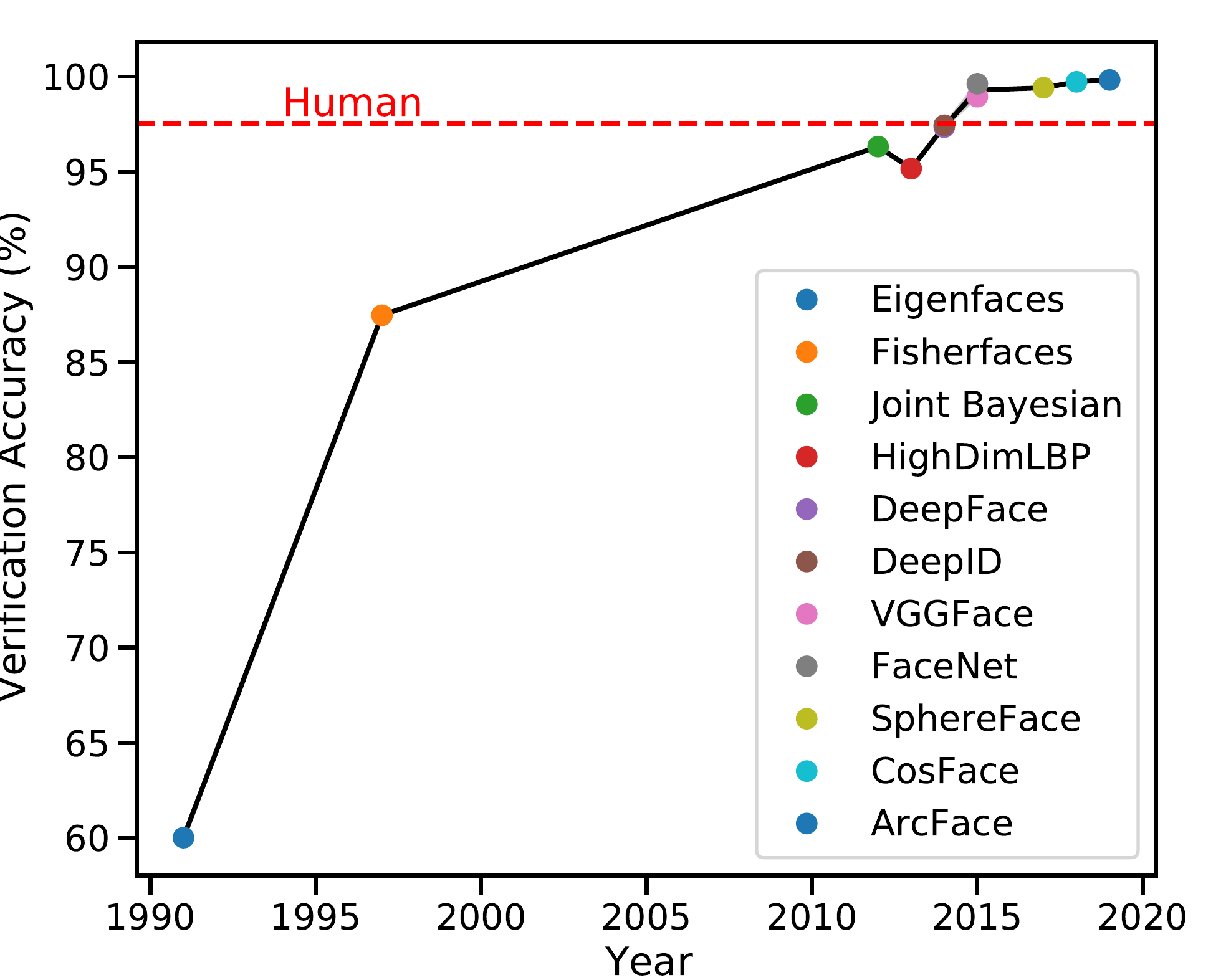}}\hfill
    \subfloat[Iris]{\includegraphics[height=1.3in]{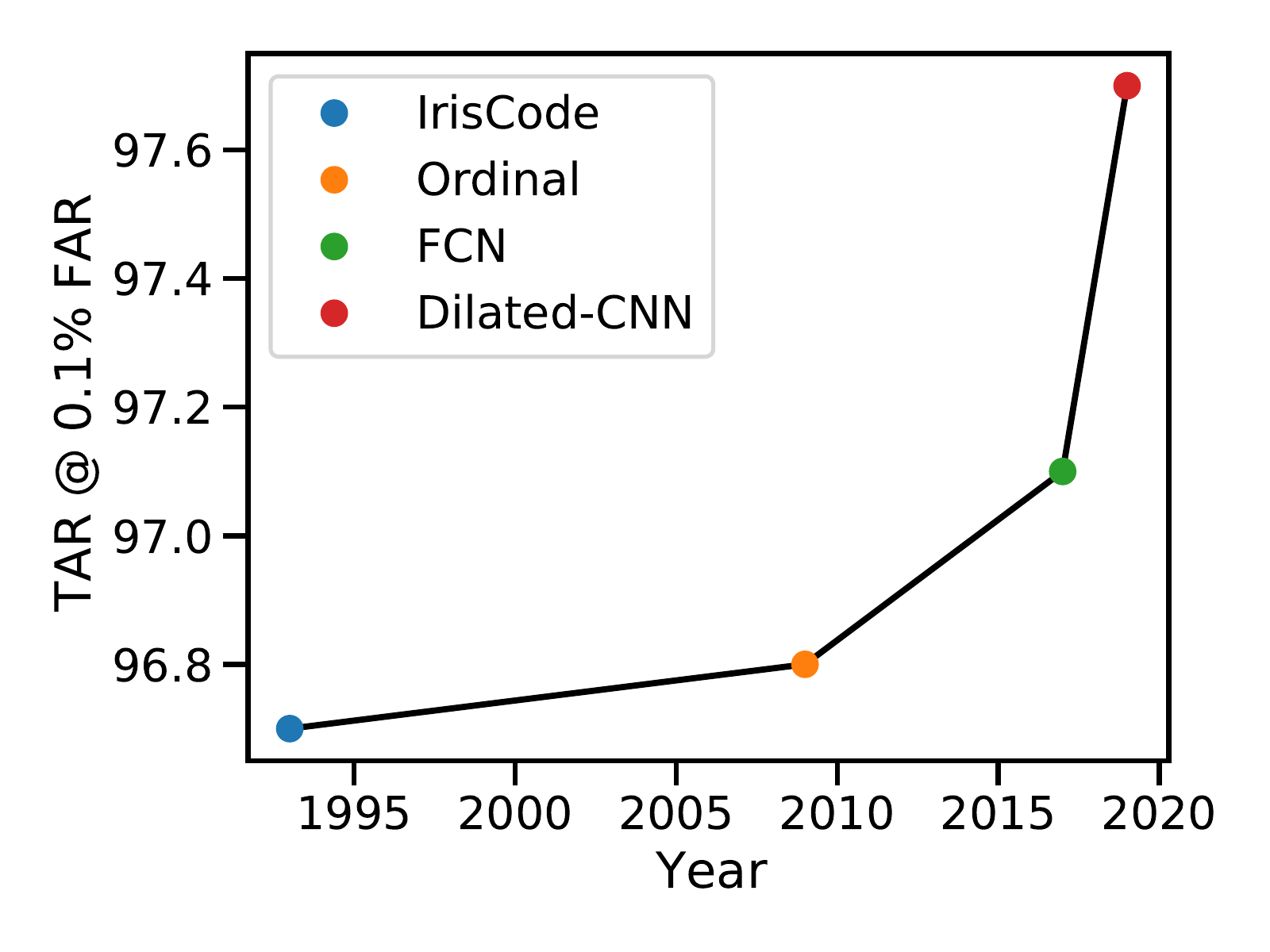}}\hfill
    \subfloat[]{\includegraphics[height=1.6in]{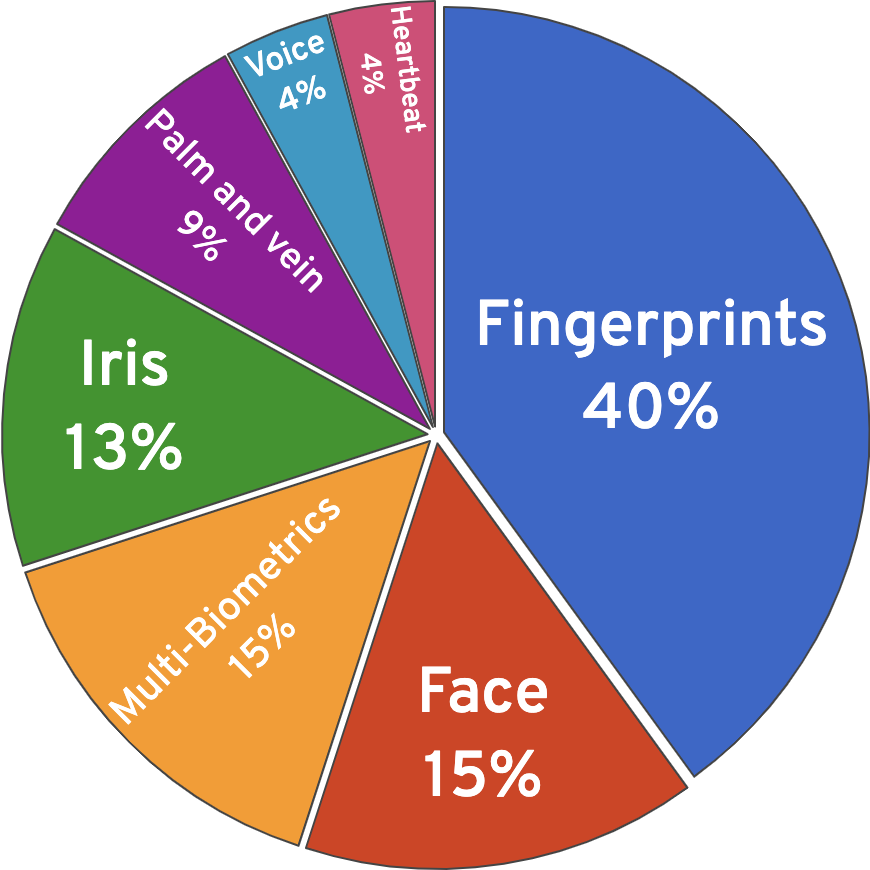}\label{fig:bio_traits}}
    \caption{Over the years, recognition rates of (a) fingerprints on the FVC Ongoing dataset~\cite{fvc}, (b) face on LFW dataset~\cite{lfw}, and (c) iris on ND-IRIS-0405~\cite{wang2019toward} have significantly improved. As a consequence, fingerprints, face, and iris recognition are widely adopted as shown in (d) compared to other biometric traits~\cite{market_share}.}
    \label{fig:perf_v_years}
\end{figure*}

\section{Recognition Performance Robustness and Scalability}

An initial prerequisite to placing trust in any recognition system is that the system is accurate. In biometric recognition systems, we expect that accuracy to be robust to various intrinsic and extrinsic noise in the input biometric signal (Figure~\ref{fig:pipeline}), and we also expect (in some cases) the system to be scalable to millions or even billions of users. In terms of accuracy, much research has been conducted since Mitchell Trauring's first paper on automated fingerprint recognition in the journal~\emph{Nature} in 1963~\cite{trauring1963automatic}. Indeed, modern day biometric recognition systems now boast accuracies in excess of human level performance (Figure~\ref{fig:perf_v_years}). However, in spite of this tremendous progress, there are still a number of situations where the biometric recognition system is not yet robust. To examine what these problems are, we first briefly lay out the inner workings of a biometric recognition system pipeline (Figure~\ref{fig:pipeline}).

A typical biometric recognition system has two stages of operation, namely, the enrollment stage (instance of the trait is captured and linked to user's credentials) and the recognition stage (a probe or query trait is compared with the enrolled trait(s)). In addition, biometric recognition systems are typically operated under one of two modes: (i) authentication (1:1 verification) and (ii) search (1:N identification). In both stages of enrollment and recognition and in both modes of authentication and search, the biometric recognition system utilizes a series of sub-modules in a systematic pipeline (Figure~\ref{fig:pipeline}). First, a biometric sensor (\eg fingerprint reader, RGB camera, or IR sensor) acquires the biometric trait (\eg fingerprint, face, or irises) of a user in digital form. Next, the digitized trait is passed to a~\emph{feature extractor} to generate a compact and salient representation (or feature set) differentiating one user from another. This representation should have high \textit{inter-class separability}, \ie different users should have very different representations. In addition, the representation should have very low \textit{intra-class variability}, \ie two representations from the same user should be very similar. The representation could be based on \textit{handcrafted features} (\eg fingerprint minutiae or iris hamming codes), learned features (\eg deep face representations), or a combination of handcrafted features with learned features (\eg through feature fusion  or by guiding deep learning methods via domain knowledge). Finally, when a user needs to be authenticated or identified, a representation extracted from the query sample can be compared to enrolled representation(s) with a matching algorithm. Breakdowns in the biometric recognition system can occur at any one of the aforementioned modules and as such, robustness and scalability must be imparted to each of them.

There are many different biometric traits, that can be utilized in conjunction with the aforementioned pipeline, however, in this paper we focus our attention on the three most popular and widely accepted traits, namely face, fingerprint and iris (Figure~\ref{fig:bio_traits}).



\subsection{Noisy Inputs}
Despite impressive recognition performance, accuracies of prevailing biometric systems are sensitive to the image acquisition conditions. For example, in unconstrained scenarios, biometric image acquisition may not be well-controlled and subjects may be non-cooperative (or even unaware). 

\begin{figure}
    \centering
    \captionsetup{font=footnotesize}
    \subfloat[]{\includegraphics[width=0.5\linewidth]{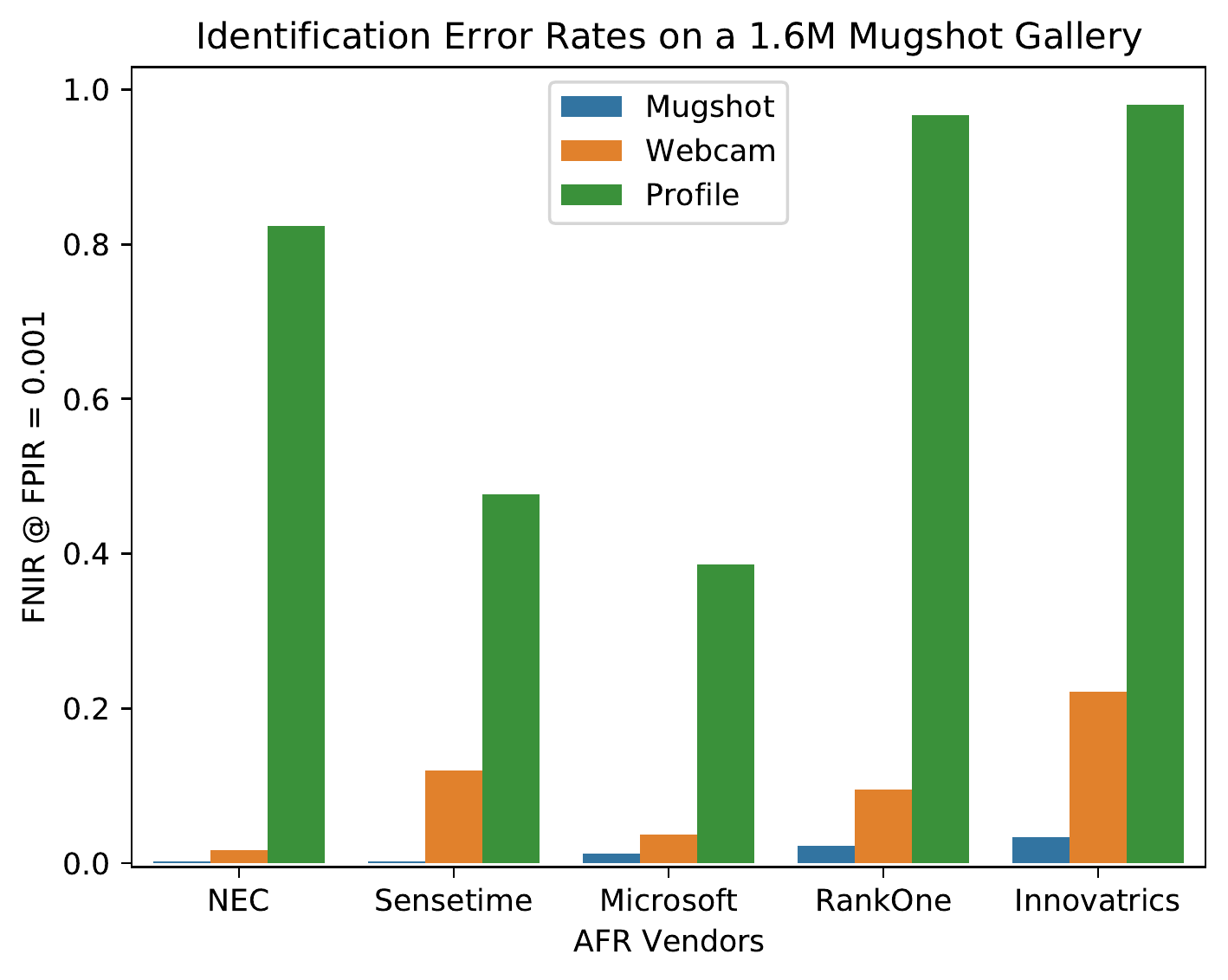}\label{fig:vendor_quality}}\hfill
    \subfloat[]{\includegraphics[width=0.5\linewidth]{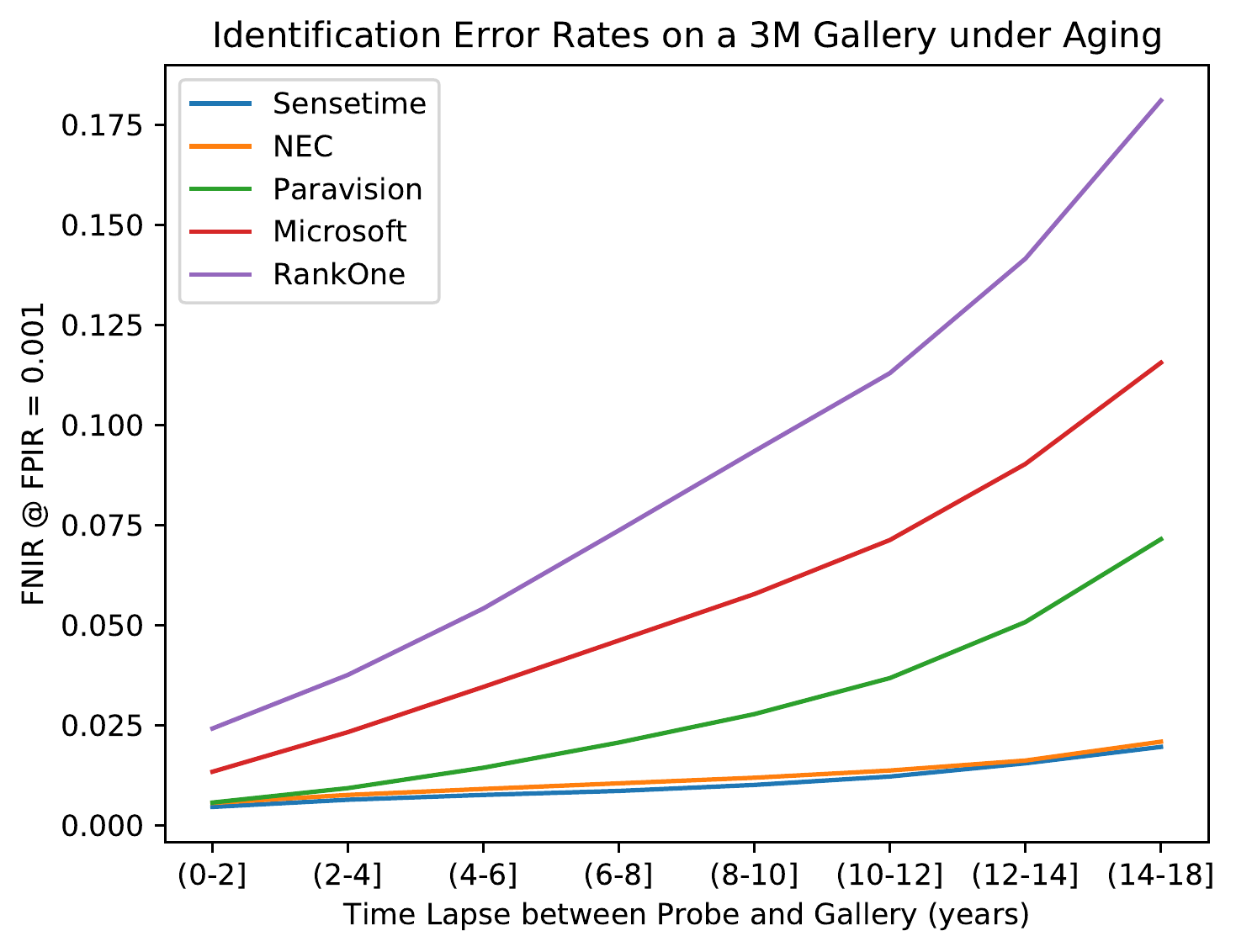}\label{fig:face_aging}}
    \caption{(a) Identification error rates of five SOTA AFR vendors when mugshots (high-quality), webcam (medium-quality), and profile (low-quality) faces are compared against a $1.6M$ mugshot dataset~\cite{grother2019frvt}. (b) Identification error rates of six SOTA AFR systems on a $3M$ mugshot dataset under aging~\cite{grother2019frvt}.}
    \label{fig:face_error_rates}
\end{figure}

\noindent \textbf{Image Quality: } The quality of a biometric image severely affects biometric recognition performance. For example, Figure~\ref{fig:vendor_quality} shows the increase in error rates when lower quality webcam and profile face photos are matched to the mugshot gallery~\cite{grother2018ongoing}. In practice, unconstrained face images are of poor image quality (such as those captured from surveillance cameras). In the case of fingerprints, images fed to fingerprint comparison algorithms may contain distortion
and motion blur due to variations in pressure applied on the sensor platen, and may have
poor contrast due to dry/wet fingers. Studies show that such degraded fingerprint images hamper recognition performance~\cite{chugh2017benchmarking, grosz2020white}. Finally, iris images which are occluded by eyelashes and eyelids can cause failures in the iris recognition system~\cite{bowyer2016handbook}. Automated person recognition performance on poor quality images is far from desirable and remains an ongoing challenge for the biometric community.

\noindent \textbf{PIE Variations: } It is now well established that accuracies of face recognition systems are adversely affected by factors including pose, illumination, expression, collectively known as~\emph{PIE}~\cite{grother2019frvt}. Fingerprints also suffer from such adverse inputs including non-linear distortion due to finger pressure and orientation and noisy backgrounds or debris~\cite{engelsma2021infant, grosz2020white, si2015detection} (\eg COTS latent fingerprint rank-1 search accuracy against a 100K gallery from an operational database is $\approx 70\%$, while rolled fingerprint rank-1 search accuracy against a gallery of 1 million fingerprints from the same database is $\approx 99\%$~\cite{cao2019end, engelsma2019learning}). Likewise iris recognition can be influenced by heavy specular reflections on the eyes~\cite{bowyer2016handbook}. While ongoing efforts in mitigating such adverse noise in biometric systems~\cite{tran2017disentangled, si2015detection, engelsma2021infant} is commendable, further research needs to be conducted for trustworthy and robust biometric systems.

\begin{figure}
    \centering
    \captionsetup{font=footnotesize}
    \includegraphics[width=\linewidth]{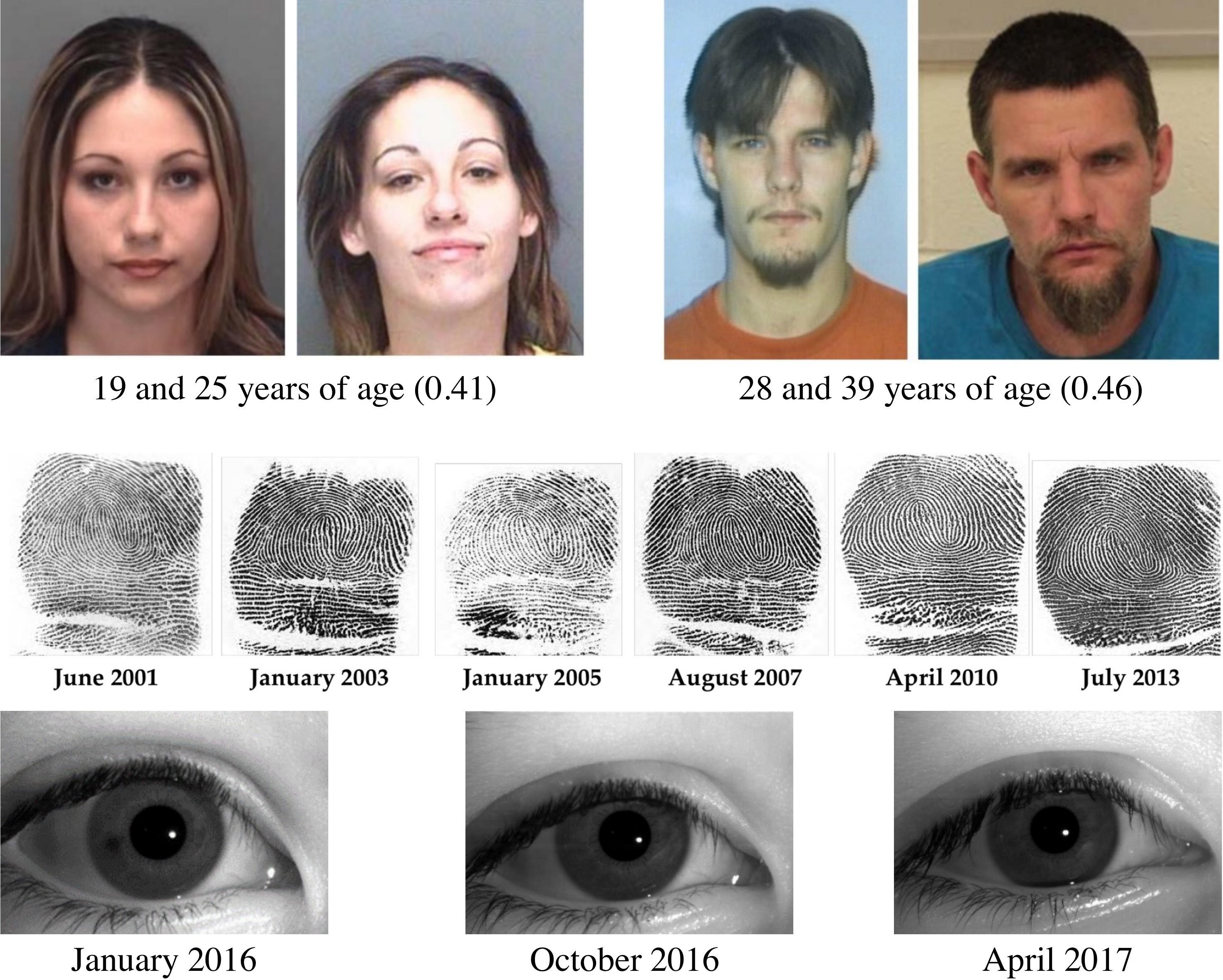}
    \caption{(Top row) Examples of low-scoring genuine face image pairs of two subjects from the PCSO longitudinal mugshot dataset~\cite{best2017longitudinal}. Ages at image acquisitions are given along with similarity scores from
COTS for each pair. COTS is a top-performing AFR vendor in the Ongoing NIST FRVT~\cite{grother2018ongoing}. (Middle row)  Fingerprint impressions from one subject in a longitudinal fingerprint dataset~\cite{engelsma2019learning}. (Bottom row) A subject's left iris images collected approximately six months apart~\cite{johnson2018longitudinal}. False rejects increase as a person's face ages, whereas, recognition performance of fingerprints and iris has been shown to be stable across large time lapses~\cite{yoon2015longitudinal, irex_vi, johnson2018longitudinal}.}
    \label{fig:longitudinal}
\end{figure}

\noindent \textbf{Aging Effects: } A considerable amount of research has been conducted to study the~\emph{permanence} of various biometric traits, \ie the trend in recognition rates as a person ages. Longitudinal studies have shown that the time gap between enrollment and gallery images have no significant impact on recognition accuracies of iris~\cite{irex_vi, johnson2018longitudinal} and fingerprint~\cite{yoon2015longitudinal} matchers. However, a human face undergoes various temporal changes, including skin texture, weight, facial hair, \etc~\cite{anatomy_face, facialstructure}.  Several studies have analyzed the extent to which facial aging affects the performance of face matchers and two major conclusions can be drawn: (i) Performance decreases with an increase in time lapse between enrollment and query image acquisitions~\cite{klare2012face, best2017longitudinal, deb2017face, grother2018ongoing}, and (ii) performance degrades more rapidly in the case of younger individuals than older individuals~\cite{grother2019frvt, deb2018longitudinal}. Figure~\ref{fig:face_aging} illustrates that state-of-the-art face matchers fail considerably when it comes to matching an enrolled child in the gallery with the corresponding probe over large time lapses (even the best face matchers begin to deteriorate after a time lapse of 10-12 years between the enrollment and probe image (Figures~\ref{fig:longitudinal} and~\ref{fig:face_aging})). Unlike other factors, face aging is intrinsic and cannot be controlled by the subject or the acquisition environment. Therefore, it is essential to enhance the longitudinal performance of biometric systems (specifically, face matchers) in order to instill trust when deployed in real-world applications such as tracing missing children~\cite{deb2020child}.


\subsection{Training Data}
Large-scale datasets have massively contributed to the improved robustness and accuracy of biometric recognition systems over the years. With the advent of deep neural networks for person recognition~\cite{engelsma2019learning, cosface, arcface}, availability of large-scale labeled dataset is paramount. For example, face recognition systems are primarily trained on 8M face images~\cite{cosface, arcface} from MS-Celeb-1M~\cite{MsCeleb} dataset, while a deep-learning-based fingerprint matcher is trained on $445$K rolled fingerprints from $38,291$ unique fingers~\cite{engelsma2019learning}. Although increasing the number of training images further could potentially improve the overall recognition performance, it is becoming exceedingly difficult to acquire large-scale face datasets with identity labels due to privacy concerns. Furthermore, large-scale datasets can introduce other challenges such as underrepresented subjects (many subjects have few images per subject)~\cite{yin2019feature}. 

Instead, an alternative approach is to collect a large set of unlabeled images to enhance the traditional supervised training setting. This can be achieved in a semi-supervised learning approach via label propagation~\cite{wan2018cost}. A different line of work explores utilizing a Graph Convolutional Network to cluster unlabeled biometric images; pseudo-labels can then be used for semi-supervised learning~\cite{yang2020learning,guo2020density,roychowdhury2020improving,zhang2020neighborhood}. Besides increasing the quantity of training data, a heterogeneous unlabeled dataset can also be introduced to augment the diversity of the prevailing labeled dataset, which has been shown to improve model generalizability to challenging and unconstrained images~\cite{shi2021boosting}.

\subsection{Scalability}

Given the success of India's Aadhaar national ID system, it would seem that biometric recognition systems have achieved a remarkable level of scalability~\cite{aadhar}. The Aadhaar system boasts over $1.3$ billion enrollees based upon de-duplication utilizing all ten fingerprints, face, and both iris images~\cite{aadhar}. However, although Aadhaar has been extremely successful in its mission to provide unique and verifiable digital identity to all, open ended questions remain in the scientific literature on the scalability of biometric recognition systems. In particular, very few evaluations exist in the literature to show how biometric recognition systems operate at a scale the size of Aadhaar (an average of $35$M biometric authentications per day\footnote{\url{https://uidai.gov.in/aadhaar_dashboard/auth_trend.php}}), the FBI's NGI program~\cite{fbi_ngi} (an average of $860$K monthly searches\footnote{\url{https://www.fbi.gov/file-repository/ngi-monthly-fact-sheet/view}}), and DHS surveillance system~\cite{dhs} (more than $350$K biometric transactions per day\footnote{\url{https://www.dhs.gov/biometrics}}). Disney Parks also employ fingerprint authentication at their entrances which encounters an average of $427$K visitors per day\footnote{\url{https://disneynews.us/disney-parks-attendance}}. If the system is not scalable and false rejects and false matches are introduced, it will cause chaos and ill-will.

Theoretically, iris recognition should be incredibly scalable~\cite{daugman2003importance}. A few studies have evaluated the search/clustering performance of face recognition against a gallery of $80$ million and $123$ million, respectively~\cite{wang2016face, otto2017clustering}. The large scale galleries were obtained by scraping photos from the web. In a similar fashion, a study was conducted in~\cite{mistry2019fingerprint} to ascertain the performance of fingerprint search algorithms against a gallery of $100$ million prints. Since there is no publicly available large-scale database for evaluating fingerprint search, the authors in~\cite{mistry2019fingerprint} first synthesize a database of 100 million fingerprints which are then used in the search evaluation. A limitation of the approach in~\cite{mistry2019fingerprint} is that a domain gap exists between synthetic fingerprints and real fingerprints such that synthetic distractors could artificially inflate the true search performance at scale. This limitation could also exist in the large scale face search studies~\cite{wang2016face, otto2017clustering} where even galleries of web scraped real data could have a domain gap with the probes from surveillance video frames. Given these challenges, and the additional increasing privacy concerns over biometric data, a very important ongoing area of research in biometrics is that of large scale synthesis. In particular, if methods can be developed to synthesize biometric images which bridge the domain gap between real and synthetic samples, better estimates on the scalability (both accuracy and speed) of biometric recognition systems can be established and consequently, biometric recognition systems can be made more trustworthy\footnote{Of course using mega-scale galleries of real data would be best for building trustworthiness, however, in practice, obtaining such datasets from legacy sources is becoming extremely difficult due to privacy concerns and/or the time and cost of collecting such an evaluation dataset.}. 

%% file: ch_security.tex
\section{Security}

Aside from the performance robustness and scalability of state-of-the-art (SOTA) biometric recognition systems discussed in the previous section, perhaps the next most important aspect of biometric recognition systems needed to solidify trustworthiness is that of their \textit{security} or their often perceived lack thereof. When talking about biometric system security, we are specifically referring to those areas of the biometric recognition system which are vulnerable to manipulation and exploitation by various malicious hackers. These ``hacks" can be carried out at each of the individual stages of the biometric recognition system as shown in Figure~\ref{fig:pipeline}. To focus our attention on the most serious threats, we dive down into a few of the major points of security concern within SOTA biometric recognition systems. In particular, security threats exist at (i) the sensor level in the form of \textit{presentation attacks}, (ii) the feature extraction module via \textit{adversarial attacks}, and (iii) the database and matching modules with template theft and subsequent \textit{template reconstruction attacks}. Each of these areas of security concern have been investigated by the biometrics research community. However, points of concern remain unaddressed, particularly with respect to their generalizability to detect new attack types and new sensors not known during their training. In this section, we define each of these attacks, discuss the state-of-the-art in mitigating against these attacks, highlight what remains unsolved, and conclude with what can be done to further enhance the security of biometric recognition systems to instill trust in their continued widespread use. 

\subsection{Presentation Attacks} In IEC 30107-1:2016(E), presentation attacks (PAs) are formally defined as:

\begin{quote}
\textit{``Presentation to the biometric data capture subsystem with the
goal of interfering with the operation of the biometric system."}
\end{quote}

\begin{figure}
    \centering
    \captionsetup{font=footnotesize}
    \includegraphics[width=\linewidth]{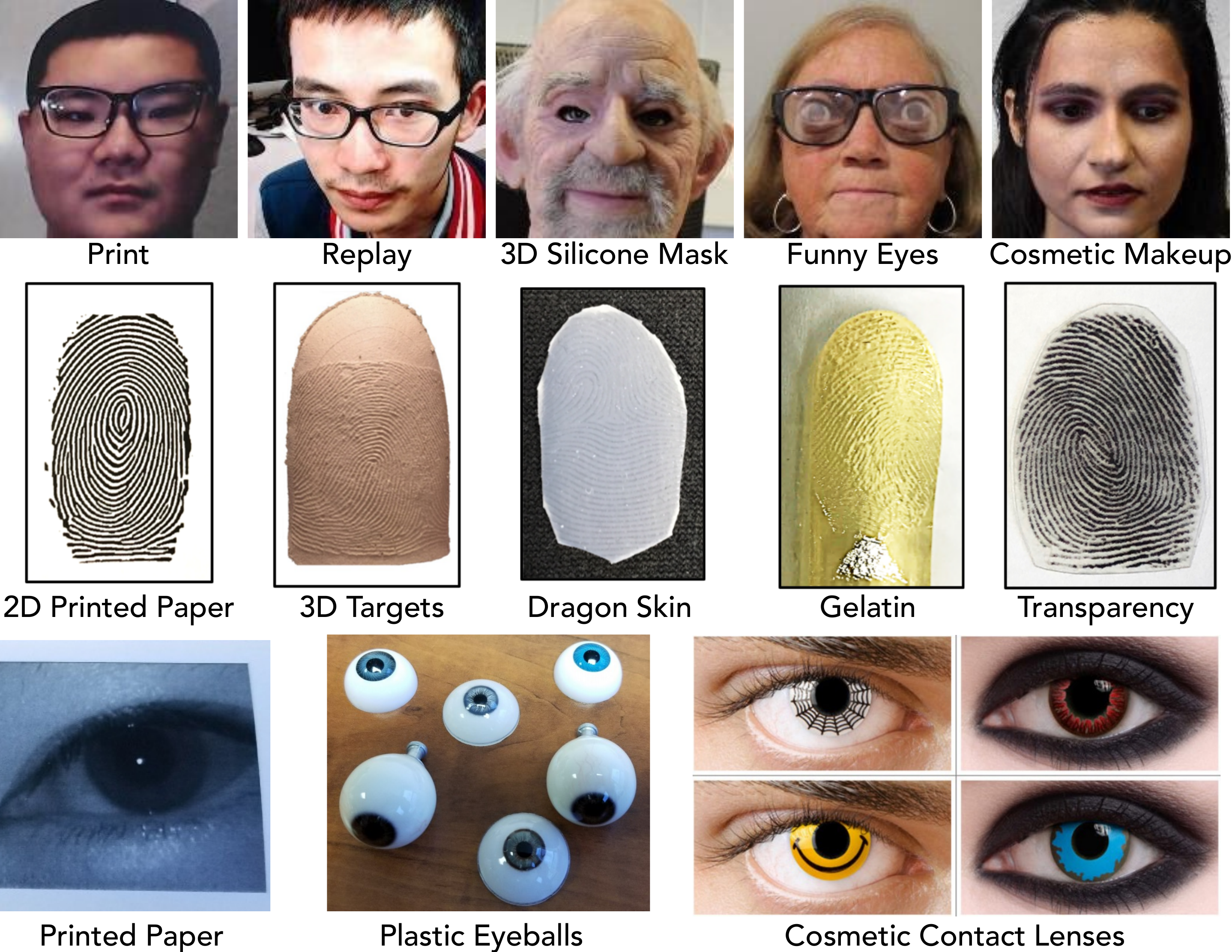}
    \caption{Examples of face (top row), fingerprint (middle row), and iris (bottom row) presentation attacks. Face spoofs are sourced from SiW-M dataset~\cite{liu2019deep}, fingerprint spoofs from~\cite{chugh2019fingerprint}, and iris spoofs from~\cite{hoffman2018convolutional}.}
    \label{fig:example_spoofs}
\end{figure}

\noindent PAs can be deployed either as an obfuscation attack (an attempt to hide one's own identity) or as an impersonation attack (an attempt to mimic someone else). For example, fingerprints could be cut or burned in an attempt to obfuscate one's identity and thus evade identification~\cite{yoon2012altered}. Alternatively, spoofs comprised of common household materials such as playdoh, wood glue, or gelatin can be used by a hacker to create an impersonation of a victim's fingerprint. More sophisticated attacks include use of high-resolution 3D~\cite{arora2016design, engelsma2018universal, schultz2021three} or 2D printing~\cite{cao2016hacking}, or cadaver fingers~\cite{schuckers2002spoofing}. In the domain of face, glasses or a mask could be used for obfuscation, while a replay on a mobile phone could be used for impersonation. Some of the well-known spoof attacks for face, fingerprint, and iris are shown in Figure~\ref{fig:example_spoofs}. In all of these examples, the attack against the biometric recognition system is carried out at the sensor level (Figure~\ref{fig:pipeline}).  

PAs have gained notoriety due to several real world examples where they have been shown to fool biometric recognition systems. For example, the German Chaos Computer Club demonstrated with ease the breaking of Apple`s TouchID already in 2013\footnote{\url{https://www.ccc.de/en/updates/2013/ccc-breaks-apple-touchid}}. Fast forward to today, fingerprint recognition systems are still being thwarted by spoof attacks with some success\footnote{\url{https://blog.talosintelligence.com/2020/04/fingerprint-research.html}}. Across biometric traits, Apple`s highly touted FaceID was compromised by a 3D mask shortly after its deployment by the Vietnamese cybersecurity firm Bkav\footnote{\url{https://www.theverge.com/2017/11/13/16642690/bkav-iphone-x-faceid-mask}}. All of these successful attacks come twenty years after early successful spoof attacks were shown in~\cite{van2000biometrical, matsumoto2002impact}. 

\begin{table}[t]
\centering
\caption{SOTA PAD Performance for Face, Fingerprint, and Iris on known (seen during training) PA types}
\label{table:benchmark_pad}
\begin{threeparttable}
\begin{tabular}{ccc}
\toprule
Trait & Competition & \specialcell{Accuracy} \\
\midrule
Fingerprint & LiveDet 2019 & 96.17\%\tnote{1} \\
\midrule
Face & \specialcell{2020 Celeb-A \\Spoof Challenge} & 100\%\tnote{2} \\
\midrule
Iris & \specialcell{2020 LivDet-Iris Challenge} & 97.82\%\tnote{1} \\
\bottomrule
\end{tabular}
\begin{tablenotes}
\item[1] Average accuracy reported in~\cite{orru2019livdet} and~\cite{das2020iris}.
\item[2] TDR @ FDR = $10^{-6}$ reported in~\cite{zhang2021celeba}.
\end{tablenotes}
\end{threeparttable}
\end{table}

The continued success in spoofing modern day biometric recognition systems is not a consequence of a lack of research into developing presentation attack detection (PAD) systems. Indeed the past couple of decades have seen a plethora of research into developing PAD systems which can automatically detect and flag a spoof attack prior to performing authentication or identification~\cite{marasco2014survey, singh2020survey, ramachandra2017presentation, galbally2014biometric, marcel2019handbook}. Typically these approaches are divided into hardware or software based approaches detecting face, fingerprint, and iris spoofs. Hardware approaches deploy additional sensors (\textit{e.g.} depth, IR cameras, multispectral illumination, \etc) to capture features which differentiate bonafide acquisitions from PAs~\cite{baldisserra2006fake, nixon2008spoof, tolosana2018towards, keilbach2018fingerprint, hussein2018fingerprint, engelsma2018raspireader, wang2013face, wang2017robust, conotter2014physiologically, zhang2011face, heusch2020deep, fang2020open}. In contrast, software based solutions extract anatomical, physiological, textural, challenge response, or deep network based features to classify an input sample as live (bonafide) or presentation attack (spoof)~\cite{marasco2014survey, george2019deep, singh2020survey, ramachandra2017presentation, galbally2014biometric, marcel2019handbook, menotti2015deep, das2020iris, yadav2020relativistic, tolosana2019biometric, yadav2021cit, hoffman2019iris+, sharma2021viability, czajka2018presentation, morales2019introduction, ferreira2019adversarial}. The culmination of these approaches can be seen in the high performances of the various algorithms submitted as part of the IARPA ODIN program\footnote{\url{https://www.iarpa.gov/index.php/research-programs/odin}} and also the public fingerprint and face liveness competitions (Table~\ref{table:benchmark_pad})~\cite{orru2019livdet, zhang2021celeba}. However, after years of rigorous research into various PAD approaches, the continued success of spoof attacks against deployed biometric recognition systems leads to the inevitable question, \textit{``What can be done to more reliably secure the biometric sensing module from spoof attacks?".} From our review of the literature, we posit that there are a few different sub-problems of biometric PAD that remain unaddressed. Solving these problems will close the spoofing loopholes remaining and will go a long way towards building trust in biometric recognition systems.

Perhaps the most significant outstanding problem with deployed PAD systems is their lack of generalization to spoofs fabricated from materials different than the spoofs that were used to train the PAD system. This problem is typically referred to as ``unseen" or ``cross" material generalization. In the domain of fingerprint recognition, multiple studies specifically showed that when a material is left out of training a state-of-the-art spoof detector and then subsequently used for evaluation, the detection accuracy drops below $10\%$~\cite{chugh2019fingerprint, engelsma2019generalizing}. Similar deterioration of unseen material detection accuracy have been observed in the face domain~\cite{liu2018learning}. In operational settings, the likelihood of a hacker using a spoof made from a novel material can be high and thus, without addressing this problem, spoof detectors remain limited in their applicability. Unfortunately, many papers continue to work on addressing ``known-material" spoof detection which already obtains nearly perfect accuracy (Table~\ref{table:benchmark_pad}) while ignoring this more challenging problem. There are a number of more recent and promising works that focus specifically on addressing the ``unseen material" and ``unseen sensor" challenge, however, the accuracy remains insufficient for field deployment~\cite{chugh2019fingerprint, engelsma2019generalizing, deb2020look, jaiswal2019ropad, gajawada2019universal, liu2019deep, kolberg2021anomaly, george2020learning, kolberg2020generalisation, george2020effectiveness, grosz2020fingerprint, mirzaalian2019effectiveness, rattani2015open, ding2016ensemble, arashloo2017anomaly, nikisins2018effectiveness, yadav2020relativistic}. Thus, we urge a stronger research push in this direction in an effort to build trustworthy biometric recognition systems.

In addition to the major vulnerability of ``unseen materials", other practical limitations of PAD systems must also be addressed. For example, many PAD systems evaluated in the literature train on one partition of a dataset captured by a particular sensor or camera, and then test on a separate partition of the same dataset (again captured by the same sensor model or camera under the same capture conditions). However, there can be a number of differences in the data distribution observed in the actual deployment scenario such as: sensor model, illumination, subject demographics, and environmental conditions. As such, models reporting near perfect accuracy on intra-dataset; intra-sensor perform quite poorly when deployed into a inter-dataset and inter-sensor scenario. We encourage PAD researchers to examine more difficult evaluation scenarios (cross dataset, cross sensor~\cite{grosz2020fingerprint, yang2019face, patel2016cross, tu2019deep, jia2020single, li2018unsupervised, wang2019improving, chugh2018fingerprint, tan2010effect}) which may be more indicative of how the PAD system will perform in the wild.

Finally, from a practical perspective, many of the PAD solutions place little emphasis on the efficiency of the PAD solution. However, many of the biometric recognition systems we use today are deployed on resource constrained devices (such as our smartphones) and as such, many of the deep learning PAD systems are impractical for real world applications. Research needs to be done to prune the parameters of the deep learning based approaches and perhaps combine deep learning approaches with simpler, faster, and lighter weight handcrafted approaches~\cite{chugh2019fingerprint, popli2021unified}.

\begin{figure}
    \centering
    \captionsetup{font=footnotesize}
    \includegraphics[width=\linewidth]{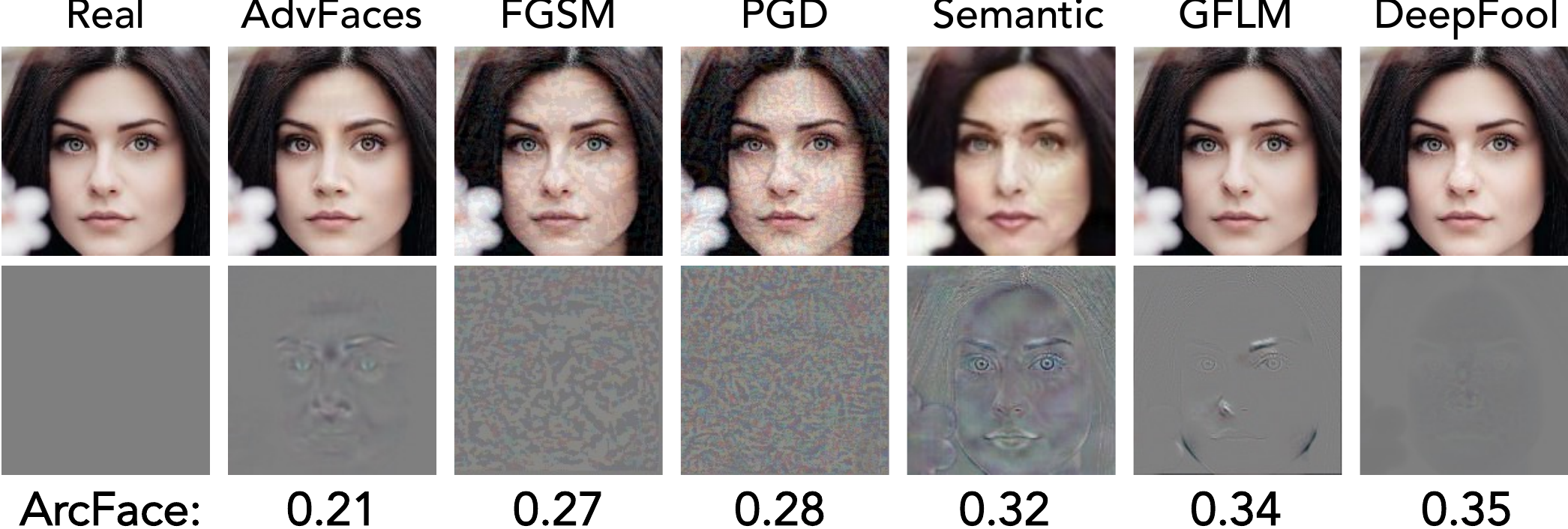}
    \caption{(Top Row) Adversarial faces synthesized via 6 adversarial attacks~\cite{faceguard}. (Bottom Row)  Corresponding adversarial perturbations
(gray indicates no change from the input). Notice the diversity in the perturbations. ArcFace match scores between adversarial image and the unaltered gallery image are given below each image. A score above
$0.36$ indicates that two faces are of the same subject. Zoom in for details.}
    \label{fig:example_adv}
\end{figure}

\subsection{Adversarial Attacks} With unrestricted access to the rapid proliferation of face images on social media platforms, such as FaceBook, SnapChat, Instagram, \etc, a community of attackers dedicate their time and efforts to digitally manipulate face images in order to evade automated face recognition (AFR) systems~\cite{faceguard}. AFR systems have been shown to be vulnerable to ``adversarial faces"\footnote{Adversarial perturbations refer to altering an input image instance
with small, human imperceptible changes in a manner that can evade CNN models~\cite{fgsm}.} resulting from perturbing an input probe~\cite{advfaces, dong, gflm, semantic_adv}. Even when the perturbations are imperceptible to the naked eye, adversarial faces can degrade the performance of numerous state-of-the-art (SOTA) AFR systems~\cite{advfaces, dong} (see Figure~\ref{fig:example_adv}). For example, face recognition performance of SOTA AFR system, ArcFace~\cite{arcface}, drops from a TAR of $99.82\%$ to $00.17\%$ at $0.1\%$ FAR on LFW dataset~\cite{lfw} when the adversarial face generator, AdvFaces~\cite{advfaces}, is encountered. Note that adversarial images are an attack on the feature extraction module of biometric recognition system (Figure~\ref{fig:pipeline}).

In contrast to face presentation attacks where the attacker needs to actively participate by wearing a mask or replaying a face photograph/video of the victim, adversarial faces do not require active participation during verification. Given the unattended nature and ``touchless" acquisition of AFR systems, an individual may maliciously enroll an adversarial image in the gallery such that at border crossing, his legitimate face image will be matched to a known and benign individual (known as an impersonation attack). An individual may also synthesize adversarial faces in order to safeguard personal privacy (\eg obfuscate automated face recognition in video conference calls~\cite{fawkes}). Also different from face presentation attacks, the adversarial perturbations are extremely subtle and directly inhibit face representations thereby making detection an extremely challenging task.

Given the growing dissemination of ``fake news" and ``deep fakes", the research community and social media platforms alike are pushing towards~\emph{defenses} against digital perturbation attacks. In order to safeguard AFR systems against these attacks, numerous defense strategies have been proposed in literature. A common defense strategy, namely~\emph{adversarial training}, is to re-train the classifier we wish to defend with perturbation attacks~\cite{fgsm, pgd, adv_train, l2l, feat_denoising}. However, adversarial training has been shown to degrade classification accuracy on real (non-adversarial) images~\cite{robustness_cost1, robustness_cost2}. In the case of face recognition, adversarial training drops the accuracy on real images in the LFW dataset~\cite{lfw} from $99.13\%$ to $98.27\%$~\cite{faceguard}. Therefore, a large number of defense mechanisms have been deployed as a pre-processing step where a binary classifier is trained to distinguish between real and perturbed faces~\cite{stochastic, artifacts, gong, grosse, li_defense, hendrycks, guo, logit_pairing, metzen, cascade,xie, efficient, uapd, smartbox, massoli, uapd, smartbox, massoli, massoli-cross, agarwal_image_transform, faceguard, goswami2019detecting, singh2020robustness}. Another pre-processing strategy, namely \emph{purification}, involves automatically removing perturbations in the input image prior to passing it to an AFR system~\cite{magnet, defense_gan, pixeldefend, self_supervised, faceguard, feat_distillation, avae}.

Similar to PAD mechanisms, an adversarial defense system also suffers from poor generalizability to perturbation types that are not encountered during its training (``unseen perturbation types")~\cite{faceguard}. In addition, employing separate pre-processing steps to detect perturbation attacks that inhibit the face feature extraction module is cumbersome and adds computational burden. Further research needs to be conducted to improve the intrinsic robustness of AFR systems to such adversarial perturbations which eliminates the need for separate detectors or purifiers.

\begin{figure}
    \centering
    \captionsetup{font=footnotesize}
    \includegraphics[width=\linewidth]{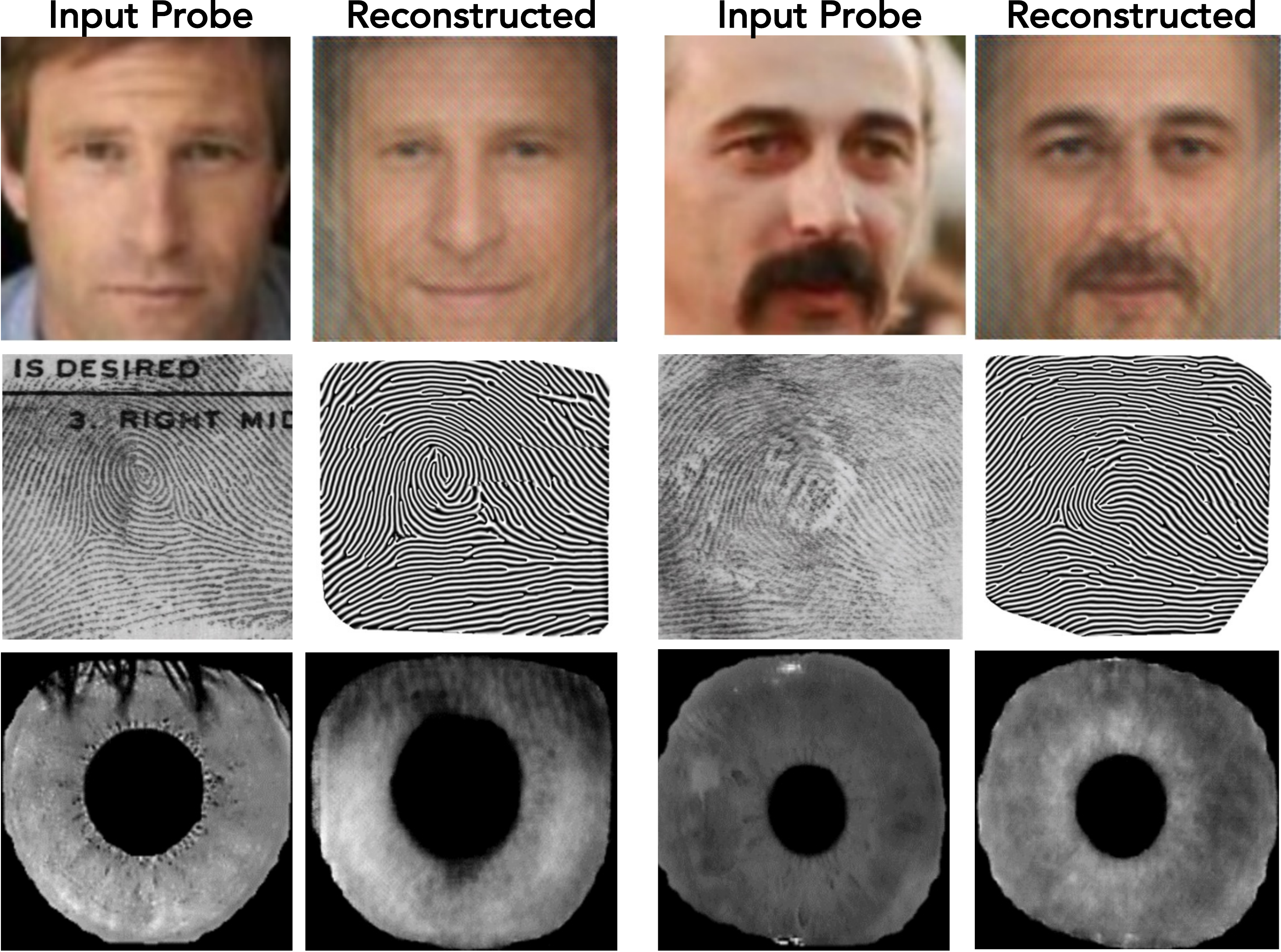}
    \caption{Two examples each of template reconstruction attacks for face~\cite{mai2018reconstruction} (top row), fingerprint~\cite{cao2014learning} (middle row) and iris~\cite{ahmad2020resist}. In all cases, the reconstruction attacks successfully match to the respective input probes.}
    \label{fig:example_template}
\end{figure}

Finally, we note that to date, adversarial attacks on the feature extraction module (Figure~\ref{fig:pipeline}) have been mostly associated with face recognition systems, since most AFRs utilize deep networks for feature extraction. In contrast, most fingerprint and iris recognition systems rely primarily on handcrafted minutiae points or iris hamming codes and are thus assumed to be safe from adversarial attacks. However, this assumption should be treated with caution as deep networks are now being explored for fingerprint and iris recognition systems as well for a number of tasks including: fixed-length representation extraction~\cite{engelsma2019learning, li2019learning, nguyen2017iris}, minutiae extraction~\cite{tang2017fingernet}, minutiae descriptor extraction~\cite{cao2019end}, spoof detection~\cite{menotti2015deep}, \etc. Presumably, any one of these deep network based fingerprint or iris algorithms could also be vulnerable to adversarial attacks. In fact, some work has been done to show that fingerprint PAD systems can be evaded by adversarial attacks~\cite{fei2020adversarial}. This is concerning since another study showed that these adversarial attacks could be converted back into a physical attack and then deployed as a successful attack on the PAD system~\cite{marrone2021fingerprint}. In addition, several successful adversarial attacks have been crafted to evade iris matchers as well~\cite{jassim2009improving, soleymani2019adversarial, soleymani2019defending, tamizhiniyan2021deepiris}.

\subsection{Template Attacks}

Finally, in addition to the security threats that exist at the sensor level in the form of spoof attacks and at the feature extraction module in the form of adversarial attacks, a very serious vulnerability of biometric recognition systems is that of limited template security. In particular, numerous studies have shown that templates extracted by biometric recognition systems (deep face representations~\cite{mai2018reconstruction}, minutiae-based fingerprint representations~\cite{feng2010fingerprint, cao2014learning}, and iriscode features~\cite{galbally2012iriscode, galbally2013iris, ahmad2020resist}) can be inverted back into the image space with high fidelity (Figure~\ref{fig:example_template}). Other studies have shown that ``soft" demographic attributes (such as age and gender) are encoded into the biometric templates~\cite{dhar2020attributes, terhorst2021comprehensive}. This is of serious concern given a number of reported breaches of databases containing biometric templates\footnote{\url{https://bit.ly/2HD83Pq}}~\footnote{\url{https://wapo.st/39PQuaT}}~\footnote{\url{https://wapo.st/2V3kHPS}}~\footnote{\url{https://bit.ly/2OQhlM3}}. Note that a template can be stolen immediately following feature extraction, as it resides in the enrollment database, or even during the matching routine if the template needs to be decrypted to perform the matching. Thus, the biometric recognition system needs to ensure that the templates remain encrypted and secured from hackers at all times.

A plethora of research has been conducted to secure biometric templates~\cite{jain2008biometric}. Some of these approaches are based upon cryptography~\cite{upmanyu2010blind} and others are pattern recognition based. For example fuzzy vault cryptosystems have been proposed for fingerprint~\cite{fuzzy1} and iris~\cite{fuzzy2} recognition. Common pattern recognition based approaches include non-invertible transformation functions~\cite{ratha2001enhancing} and cancelable biometrics~\cite{patel2015cancelable}. Another approach that has been tried is to bind a secret key with a biometric template~\cite{pr1, pr2}. Finally, techniques based on deep networks~\cite{mai2020secureface} and representation geometry~\cite{kim2021ironmask} have been proposed. All of these approaches are limited in that they trade off the recognition accuracy of a biometric system for the enhanced security.

A more recent development in biometric template protection is that of homomorphic encryption (HE) systems~\cite{he1, he2, he3, he4, he5}. Homomorphic encryption enables doing basic arithmetic operations directly in the encrypted domain. Because of this, the primary benefit of using HE is that it can protect the template as it resides in the database, and also while it is being compared (assuming the matching function can be reduced to arithmetic operations of addition and multiplication, \textit{e.g.} the cosine similarity between two face representations). The limitation of HE systems is that it is computationally expensive, especially Fully HE systems which allow for both addition \textit{and} multiplication operations directly in the encrypted domain. Work has been done to alleviate the computational burden of FHE for biometric matching~\cite{fhe1, fhe2, fhe3}, however, research remains to further speed up this encrypted matching process (\eg~the work in~\cite{fhe3} showed encrypted fingerprint search against $100$ million gallery in $500$ seconds, a $275\times$ speedup over SOTA; the same search in the unencrypted domain would take $10$ seconds~\cite{engelsma2019learning}).

Generally speaking, all of the methods that attempt to better protect the biometric template, seek a compromise along multiple axes of speed, memory, accuracy, and security. Research must continue to minimize the trade-offs and sacrifices that occur in any one of these dimensions. Ideally, a trustworthy biometrics recognition system would secure the template, at all times, while sacrificing very little along any of these axes.

\subsection{Unifying Security Efforts}
As an addendum on the security efforts across sensing, feature extraction, and matching modules, we note that prevailing research efforts focus on mitigating~\emph{one} of the three attack categories at a time: (i) presentation attacks, (ii) adversarial attacks, and (iii) template attacks. Since the exact type of biometric attack may not be known~\emph{a priori}, researchers are encouraged to design~\emph{generalizable} detectors that can defend biometric systems against any of the three attack categories~\cite{unifad} (\eg in an enrollment scenario, a single detector could quickly check for live vs. spoof, adversarial perturbations, and reconstruction attacks). Such systems will alleviate the computational burden of securing the entire biometric recognition pipeline. 

%% file: ch_interpret.tex
\begin{figure}
    \centering
    \captionsetup{font=footnotesize}
    \includegraphics[width=\linewidth]{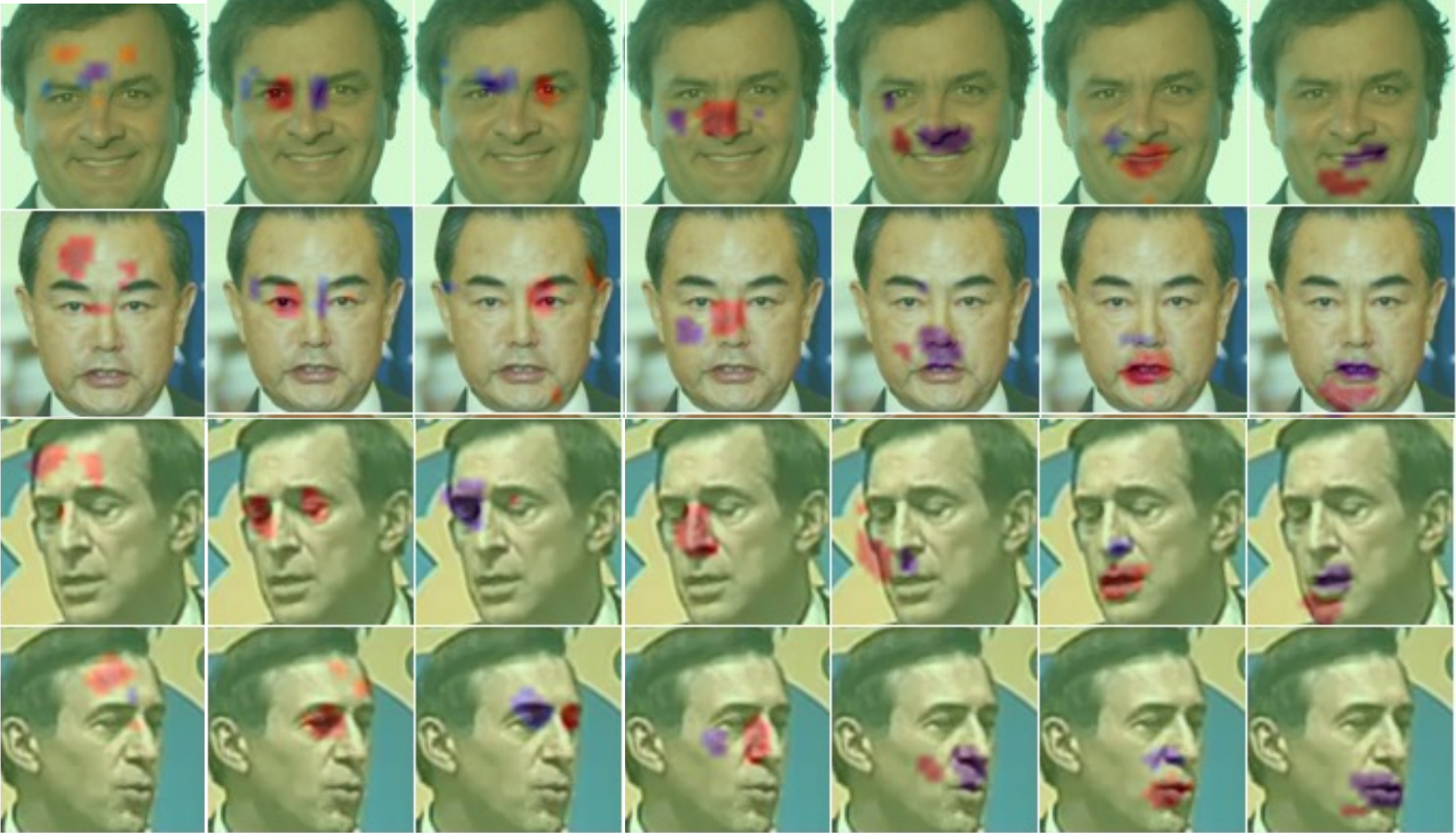}
    \caption{Visualization of filter response ``heat maps" of 7 different filters from an~\emph{interpretable face recognition system}~\cite{yin2019towards} on face images from different subjects (Top 2 rows) and the same subject (Bottom 2 rows). The positive and negative responses are shown as two colors within each image. Note the high consistency of response locations across subjects and across poses~\cite{yin2019towards}.}
    \label{fig:explain_face}
\end{figure}

\section{Explainability and Interpretability}

In addition to being accurate and secure, a trustworthy biometric recognition system should also have a certain degree of interpretability such that system designers and agency deploying the system can understand why a decision is made and adjust the system's decision if needed (\ie by inserting a human in the loop). Interpretability is also important in courts of law, where fingerprint and face evidence could be used to convict a person~\cite{innocent2019, committee2009strengthening}. For example, if we are using a face recognition system's prediction to identify someone as a criminal, we would like to understand why the system thinks the probe and gallery faces appear similar to prevent potential false convictions or false acquittals\footnote{Interpretability can also greatly aid the judge and the jury in cases where both the prosecution and the defense present conflicting recognition results based on their own proprietary black-boxes.}. However, most deep neural network based models, utilized for face recognition, serve as black boxes that give final decisions on probe samples directly via millions of learned parameters. 

To better impart credibility and interpretability to these black box systems, many methods have been proposed in the broader computer vision and machine learning community. One popular direction of research is visualizing the features that are learned in the model~\cite{zeiler2014visualizing,mahendran2015understanding,yosinski2015understanding,dosovitskiy2016inverting,olah2017feature}. Others focus on the attribution of the decision, either by finding the features~\cite{sundararajan2017axiomatic,lundberg2017unified,shrikumar2017learning} or the local regions in images~\cite{simonyan2013deep,fong2017interpretable,zhou2016learning,selvaraju2017grad,smilkov2017smoothgrad} that lead to the final decision. Although the feature visualization methods could be directly applied to the feature extraction module of biometric systems, the attribution methods may be better geared towards classification models used in biometrics (\ie PAD algorithms) since the goal is to interpret a final classification decision.

\begin{figure}
    \centering
    \captionsetup{font=footnotesize}
    \includegraphics[width=\linewidth]{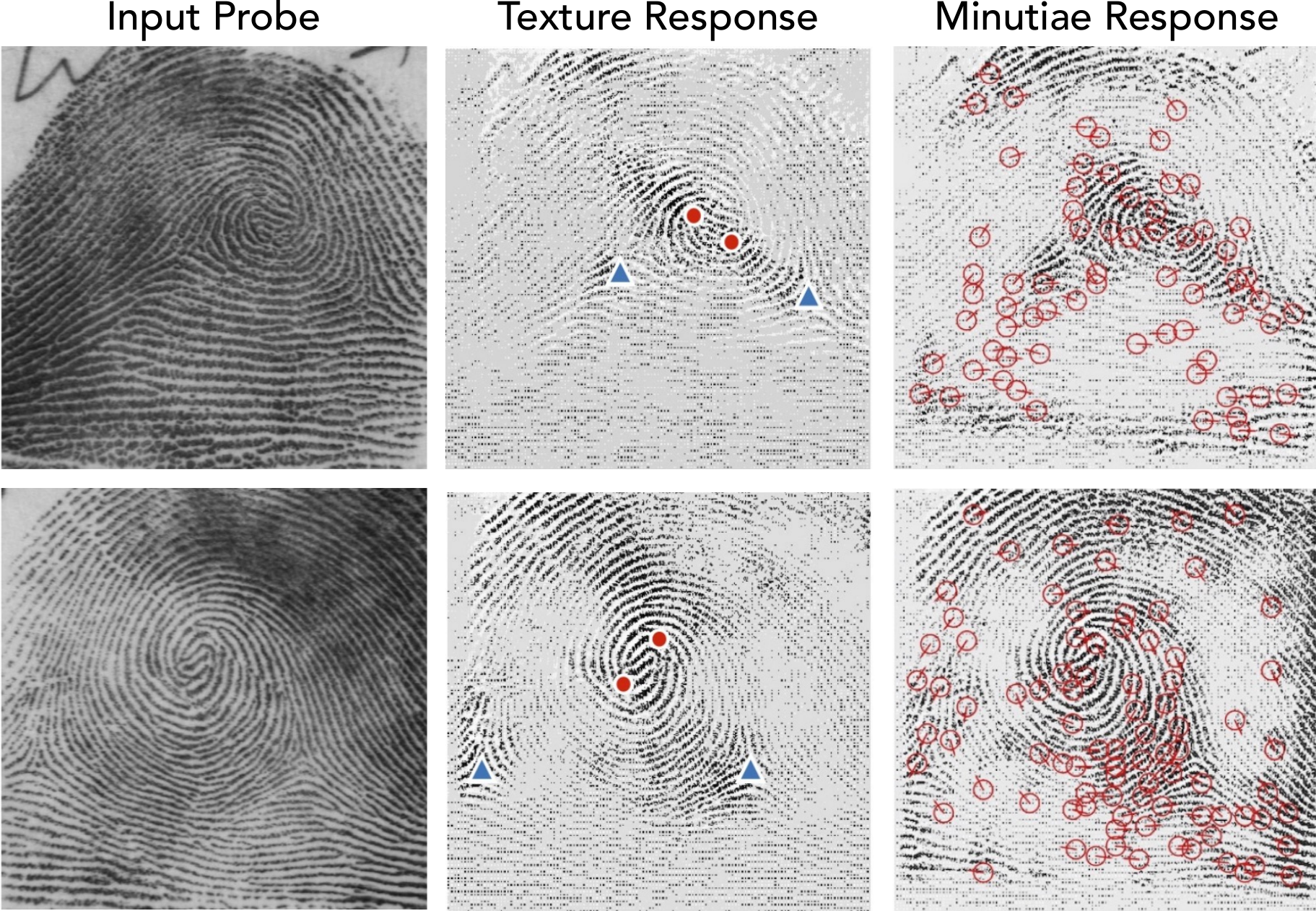}
    \caption{Visualization of filter responses from texture and minutiae branches on two fingerprint images via DeepPrint, a CNN-based fingerprint matcher~\cite{engelsma2019learning}. This shows us that the network learns to extract features related to areas of the fingerprint we know are discriminative (minutiae and singularity points).}
    \label{fig:explain_fpt}
\end{figure}

Within the biometric modality of face specifically, a few studies have attempted to understand how features are learned and used to compare faces. For example, Yin~\etal~\cite{yin2019towards} propose to constrain the learning stage such that features are directly related to different areas of the face. Once the models are trained, saliency maps can be used to visualize which part of the face a filter is looking at (see Figure~\ref{fig:explain_face}). Experimental results also show that in addition to imparting spatial interpretability, regularizing the spatial diversity of the features enables the model to become more robust to occlusion. A drawback of Yin's method is that a model needs to be re-trained to obtain such interpretability (\ie interpretability can not be extracted from prevailing commodity AFR systems). In response, Stylianou~\etal~\cite{stylianou2019visualizing} propose a model-agnostic method that visualizes the salient areas that contribute to the similarity between a pair of faces. It is observed that models are indeed focusing on the entire face, but they were not able to provide more fine-grained details, such as which part of the face is contributing to the similarity/dissimilarity between a pair of faces. Therefore, we believe this problem of similarity attribution remains a meaningful yet unsolved problem for future research. 

In another line of research, the studies in~\cite{dhar2020attributes} and~\cite{terhorst2021comprehensive} provide interpretability to AFRs by studying how facial attributes are encoded in the deep neural network. In particular, the authors in~\cite{dhar2020attributes} chose four common attributes, namely identity, age, gender and face angle (yaw), and estimated their correlation with face representations. They found that compared to low-level features, high-level deep face representations tend to be more correlated with identity and age while less correlated with gender and face angle. In a similar line of research, the authors in~\cite{terhorst2021comprehensive} examined the effect of 47 high-level attributes on face recognition performance. They observed that many nuisance factors such as accessories,
hair-styles and colors, face shapes, or facial anomalies influenced the face recognition performance.

\begin{figure}
    \centering
    \captionsetup{font=footnotesize}
    \subfloat[Face Spoof Regions]{\includegraphics[width=\linewidth]{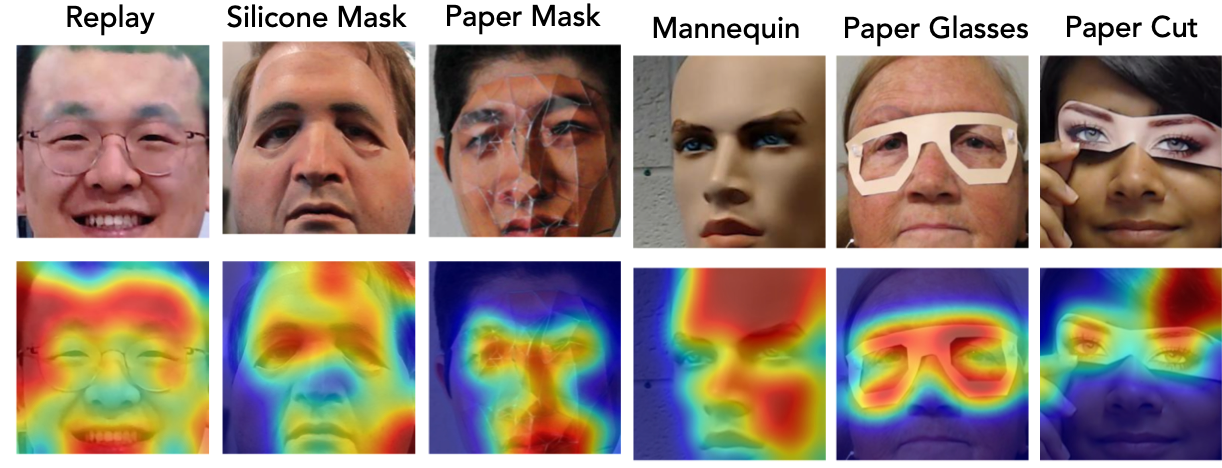}}\\
    \subfloat[Fingerprint Spoof Regions]{\includegraphics[width=0.6\linewidth]{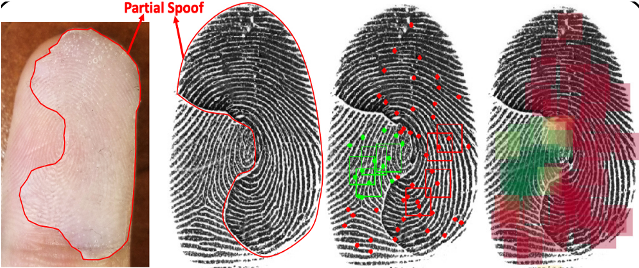}}\hfill
    \subfloat[Iris Spoof Regions]{\includegraphics[width=0.4\linewidth]{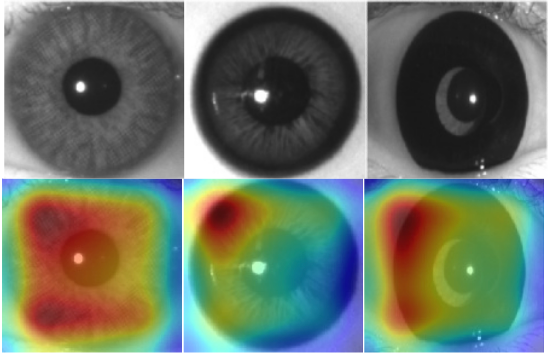}}
    \caption{Visualizing (a) face spoof~\cite{deb2020look}, (b) fingerprint spoof~\cite{chugh2018fingerprint}, and (c) iris spoofs~\cite{chenexplainable} regions. Blue regions in (a, c) indicate bona fide regions while red regions denote spoofs. Red regions in (b) indicate likely fingerprint spoof or fingerprint alteration.}
    \label{fig:explain_spoof}
\end{figure}

A more recent direction in interpreting and improving AFRs is through uncertainty estimation. For example, Shi~\etal proposed in~\cite{shi2019probabilistic} to represent each input face as a distributional representation in the feature space (rather than a single point or feature vector), where the variance of the distributions represent the uncertainty of the corresponding features. Besides improving the face recognition performance, they showed that the feature uncertainty could also be used to visualize the perception of the model about the input.

While all of the aforementioned methods have certainly helped impart more interpretability to AFRs, there is still much we do not yet know and understand about what information about the input image is being encoded into deep face representations. Having a better understanding of what these encodings are comprised of could help address biasness and other failures in the AFR system. Interpretability also needs to be extended to other modules of the face recognition pipeline (such as spoof detection) where nuisance factors could potentially cause a spoof face to be misclassified as a live face. Therefore, we posit that more work in this area remains to be done in an effort to build trustworthy biometric recognition systems.

We note that much of the interpretability concerns mentioned thus far have been centered around face recognition systems. This is because nearly all face recognition systems employ the use of ``black-box" deep networks for encoding and matching. However, as per our earlier discussion on adversarial attacks, deep networks are now being increasingly used for fingerprint and iris recognition systems as well. Thus several studies have begun to more carefully discuss interpretability of deep networks deployed for various tasks within the fingerprint and iris recognition pipeline. For example, the authors in~\cite{engelsma2019learning}, utilize the feature attribution method from~\cite{zeiler2014visualizing} to visualize the features being learned by a deep network for fixed-length fingerprint representation extraction (Figure~\ref{fig:explain_fpt}). They conclude that the network is able to automatically learn areas in the fingerprint image that are already deemed highly discriminative (singularity points, and minutiae points). A similar observation was made in~\cite{chowdhury2020can}. Akin to AFR systems, as fingerprint matchers begin to rely more on deep networks, further research needs to be conducted to ensure the interpretability of their decisions. Aside from interpreting deep learning based methods in the domain of fingerprint, some other studies have tested minutiae-based fingerprint matchers to determine which source of noise contributes the most to the final fingerprint recognition decision~\cite{grosz2020white, chugh2017benchmarking}.

Finally, interpretability is not limited to the feature extraction and matching modules of the biometric recognition pipeline. For instance, researchers working on face, fingerprint, and iris spoof detection modules have also begun examining more closely the types of features that a deep network uses to differentiate a live biometric sample from a spoof~\cite{deb2020look, chugh2019fingerprint, liu2020disentangling, liu2020physics, chenexplainable, sharma2020d}. This is especially important in order to prevent wrongfully denying access to genuine subjects. For example, in the event that a person is flagged for attempting to spoof a biometric system, the PAD system should visualize which regions of the biometric sample consists of a spoof to further aid a human operator doing a manual inspection; a global ``spoofness score" alone may not be sufficient for a human operator to interpret the network’s decision (see Figure~\ref{fig:explain_spoof}). 

%% file: ch_bias.tex
\begin{figure}
    \centering
    \captionsetup{font=footnotesize}
    \includegraphics[width=.475\linewidth]{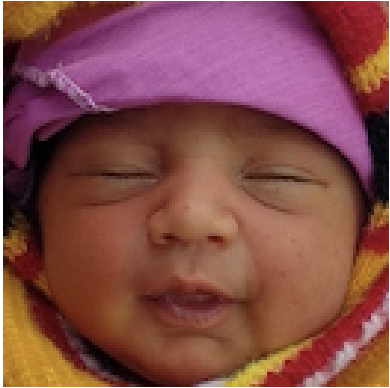}\hfill
    \includegraphics[width=.475\linewidth]{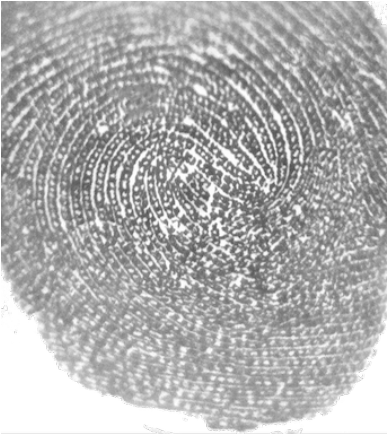}
    \caption{Face image and corresponding thumb-print of a 1-week old infant~\cite{engelsma2021infant}. The thumb-print was captured and matched using a custom 1,900 ppi reader and accompanying high-resolution matcher, since the standard 500 ppi COTS readers and matchers do not have sufficient resolution to capture and match an infant's fingerprints. }
    \label{fig:bias-infant}
\end{figure}

\begin{figure}
    \centering
    \captionsetup{font=footnotesize}
    \includegraphics[width=\linewidth]{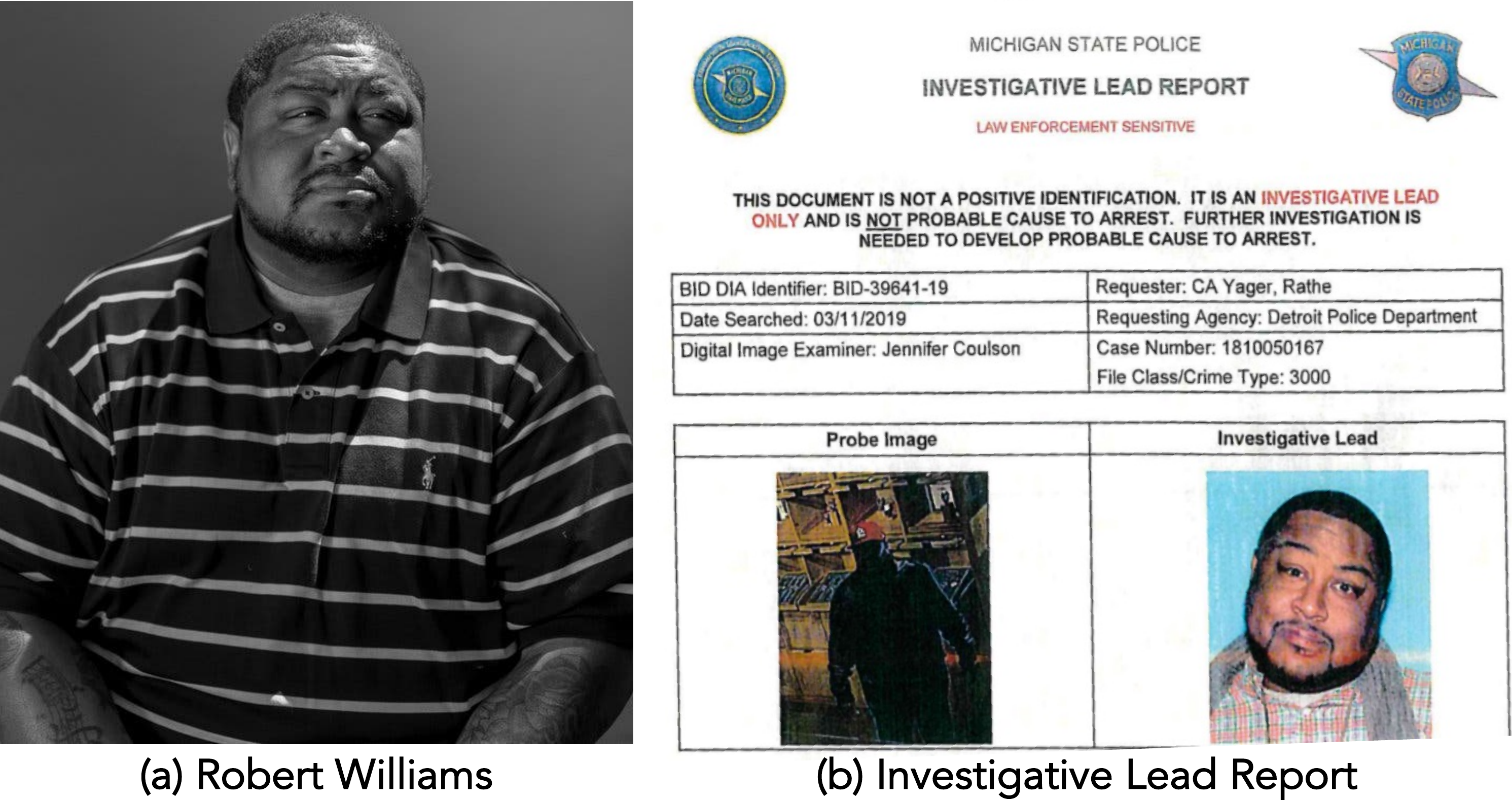}
    \caption{Face recognition system wrongfully identified (a) Robert Williams when the CCTV frame in (b) is searched against a 49M gallery. On arrest, Williams responds, \emph{``This is not me. You think all Black men look alike?”}~\cite{nyt_bias}. Relying on automated person recognition alone may lead to a lack of trust in the eyes of policymakers and citizens alike. Therefore, it is imperative to have ``humans-in-the-loop" where human examiners can verify the decisions made by biometric systems. In addition, biometric systems should also have a ``reject" option instead of  match/non-match binary decisions.}
    \label{fig:bias-nyt}
\end{figure}

\section{Demographic Bias and Fairness}
\begin{figure}[t!]
    \centering
    \captionsetup{font=footnotesize}
    \includegraphics[width=0.97\linewidth]{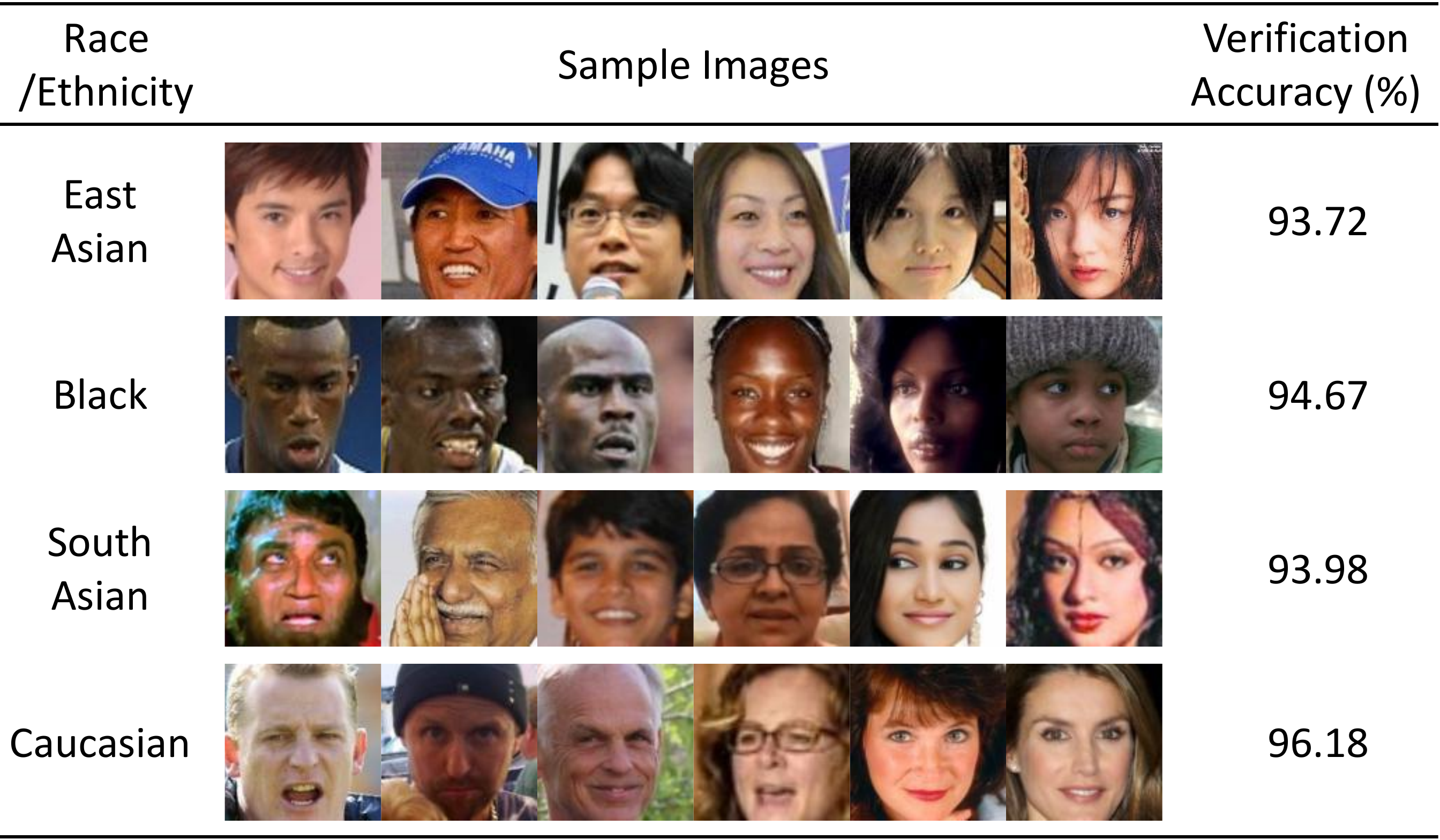}
    \caption{Face Verification performance by ArcFace~\cite{arcface} on each race/ethnicity cohort in RFW dataset~\cite{wang2019racial}.}
    \label{fig:bias-face-acc}
\end{figure}

Another issue of trust with biometric recognition systems that has more recently been brought to light in mainstream media is that of biased performance against certain demographic groups~\cite{howard2019effect, klare2012face, grother2019frvt, gender_shades}, referred to as \emph{demographic bias in biometrics}. 
When a biometric system is defined to be demographically biased, it algorithmically provides higher recognition performance for users within a subset of demographics and lower performance in other demographic groups (see Figure~\ref{fig:bias-face-acc}). In fact, all 106 face recognition algorithms (from academia and industry alike) that were submitted to the NIST FRVT~\cite{grother2019frvt} exhibit different levels of biased performances based on gender, race, and age groups of a mugshot dataset. Similar bias issues in AFRs were reported by earlier studies on demographic attribute estimation~\cite{news-mit}. It should be acknowledged that the demographic bias shown by the best performing commodity AFR systems in the NIST FRVT on~\emph{mugshot faces} is less than $1.0\%$ across the four groups: Black Male, Black Female, White Male and While Female~\cite{grother2019frvt}. Furthermore, every top-tier AFR system studied in NIST FRVT Ongoing is most accurate on Black Males~\cite{grother2018ongoing}. It should also be noted that the extent of bias across different demographic cohorts cannot be precisely known until proper ground truth adjudication can be done on large scale datasets such as that used in the NIST FRVT (where the level of performance on different demographic groups flipped before and after manual ground truth adjudication)~\cite{grother2019frvt}. 

Since facial regions contain rich information of demographic attributes, most studies on bias are focused on face-based biometrics~\cite{klare2012face, howard2019effect, grother2019frvt, gong2020jointly, deb2017face}. However, several studies have also investigated the bias factor of age or aging in other biometric modalities (fingerprint~\cite{yoon2015longitudinal, jain2016fingerprint, j2019infant, engelsma2021infant}, or iris~\cite{fang2021demographic}). A consistent finding of bias in face recognition across studies in~\cite{klare2012face, best2017longitudinal, deb2017face} is that the recognition performance is worse for female cohorts (possibly due to the use of cosmetics). The studies of~\cite{buolamwini2018gender, raji2019actionable} showed a significant attribute estimation accuracy impact based on age, gender and race.  In the domain of fingerprint,~\cite{yoon2015longitudinal} indicated a non-trivial impact of age on genuine match scores. 

\noindent \textbf{Biometrics for Lifetime: } Multiple studies have shown the extreme difficulty in performing biometric recognition on the most vulnerable amongst us, namely infants and young children (Figure~\ref{fig:bias-infant})~\cite{jain2016fingerprint, j2019infant, engelsma2021infant}.

Most of the aforementioned studies address algorithmic demographic bias, however, we also highlight the role that biometric sensors can play in biasness. For instance, matching fingerprints from different sensors is a challenging problem~\cite{ross2004biometric}.
In the case of iris recognition, brown-eyed individuals are more susceptible to sensor issues and therefore, near infra-red sensors are adopted instead of RGB cameras. Finally, a study on AFRs showed that ``the magnitude of measured demographic effects depends on image acquisition" ~\cite{cook2019demographic}. 

Deploying biased systems could come with significant consequences, especially against those whom the system does not perform as well on, \eg being unjustly incarcerated or denial of bail or parole~\cite{osoba2017intelligence, washington2018argue, garvie2016perpetual, o2016weapons} (see Figure~\ref{fig:bias-nyt}).
Therefore, it is crucial to estimate and mitigate demographic bias in biometric recognition systems. Such systems should show no statistically significant difference on the performance amongst different demographic groups of individuals. At the same time, the overall accuracy of the system should not be compromised, ideally.

\begin{figure*}
    \centering
    \captionsetup{font=footnotesize}
    \includegraphics[width=\linewidth]{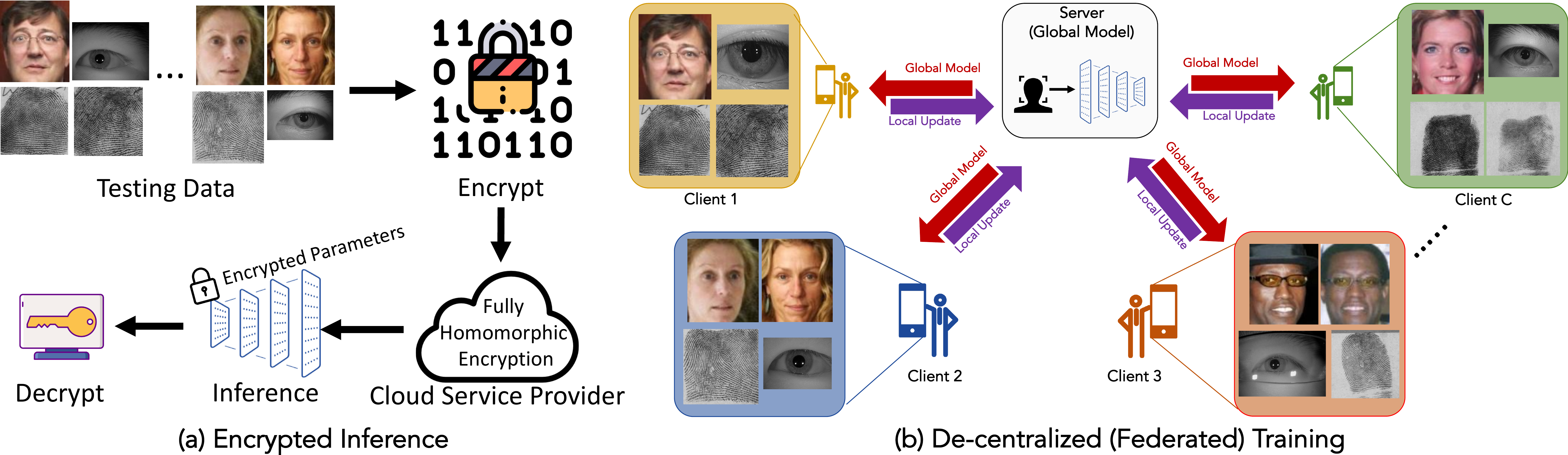}
    \caption{Two potential methods of imparting user privacy to biometric systems. A number of studies explore using (a) homomorphic encryption to perform inference on encrypted data with encrypted model parameters~\cite{nalini1, nalini2, nalini3, gilad2016cryptonets, brutzkus2019low, yonetani2017privacy}. Another approach to privacy involves training the biometric system in a (b) de-centralized (federated) manner where model parameters are shared between clients and server~\cite{fedface}.}
    \label{fig:overview_privacy}
\end{figure*}


To mitigate bias in biometric recognition systems, a simple question must first be answered: \emph{What factors lead to bias in biometric recognition systems?} The answer to that question is multi-faceted.
First of all, many state-of-the-art biometric recognition systems are based on deep networks, which rely on large training datasets. These training datasets of human subjects are often biased towards certain demographic cohorts. Secondly, the implementation of biometric recognition systems can be statistically biased during the learning process, for example, by parameter optimization and regularization. For example, a representation extraction network undergoing training is typically trying to satisfy training samples on the average case while potentially placing less weight on under-represented samples leading to biasness. 
Finally, the fourth factor is what is referred to as \emph{intrinsic bias}, a notion first introduced by~\cite{klare2012face}, stating that subjects in certain demographic groups are inherently more difficult to be recognized. 

Given the various sources of bias mentioned above, bias mitigation requires special attention on both data sampling and algorithm design. 
Early studies on dataset-induced bias include data re-sampling methods (oversampling or undersampling images of certain demographics)~\cite{drummond2003c4, chawla2002smote, mullick2019generative}. Data re-sampling is limited in that useful, diverse information is discarded. 
Therefore, rather than re-weight the sample distribution in the training set, later studies tackle bias by re-weighting the loss values in objective functions~\cite{cao2019learning, cui2019class}, also called \emph{cost-sensitive learning}, based on a sample's demographic cohort.

The aforementioned works do not take into account the correlation between demographics and identity. As such,~\cite{gong2020jointly} proposes a framework to jointly learn unbiased representations for both the identity and demographic attributes by disentangling them. The impact of bias is mitigated by removing sensitive information (demographics or identity) from each component of the disentangled representation. A limitation of~\cite{gong2020jointly}, is that the overall recognition performance declines. To be practical, algorithms mitigating bias in face recognition should also maintain the overall recognition accuracy. 
To address this challenge, Wang~\etal~\cite{wang2020mitigating} propose an adaptive margin for faces in each demographic group. 
Another approach proposed by~\cite{gong2021mitigating} adapts the network operations by employing dynamic convolutional kernels and attention maps based on the demographic group. 
Both~\cite{wang2020mitigating} and~\cite{gong2021mitigating} manage to improve the performance on under-represented groups while better maintaining the overall accuracy. 

Despite recent progress in mitigating demographic bias, this issue has not been completely rectified and still demands further research, especially given the fact that a variety of factors could lead to bias other than the predefined demographic groups that most studies assume. Existing studies need to make sure that overall system accuracy is not compromised via bias reduction. 
Furthermore, since the majority of the existing studies are concentrated on bias mitigation for face-based biometrics, there is an urgent need for research on other biometric modalities (e.g., fingerprint~\cite{marasco2019biases}). Finally, biasness research should also be conducted on algorithms other than the recognition system (\ie the PAD modules, where biasness could inconvenience users of certain demographics unfairly). Biased biometric recognition systems create an element of mistrust in the general public and as such, removing this bias is a critical step on the path towards trustworthy biometric recognition systems.

%% file: ch_privacy.tex
\section{Privacy}

A final key area of biometric recognition systems that we posit is necessary in order to build trust is that of user privacy. Note, we explicitly differentiate between \textit{security} (such as the template security previously discussed) and \textit{privacy}. While security is aimed at addressing attacks on the biometric recognition system with the goal of interference, privacy does not necessarily entail an attack. Rather it entails the respect and confidentiality of an individual's personal identifying information (PII) or data as well as transparency surrounding its use and storage.

A number of high profile laws have been enacted to better ensure privacy. In 2008, the Illinois legislature unanimously passed the Biometric Information Privacy Act (``BIPA"), based on efforts by the ACLU\footnote{\url{https://www.aclu-il.org/en/campaigns/biometric-information-privacy-act-bipa}}. The Illinois law enables individuals a better control of their own biometric data and prohibits private companies from collecting it unless they:

\begin{itemize}
    \item Inform the person in writing of what data is being collected or stored.
    \item Inform the person in writing of the specific purpose and length of time for which the data will be collected, stored and used.
    \item Obtain the person’s written consent.
\end{itemize}

\noindent Likewise, in 2016 the GDPR~\cite{eu_gdpr} (General Data Protection Regulation) passed the European Parliament. The GDPR defined personal data (to include biometric data) as: \textit{``... any information that relates to an individual who can be directly or indirectly identified ... including biometric data, ...".}. The GDPR further laid out strict guidelines for processing data:

\begin{itemize}
    \item Lawfulness, fairness and transparency — Processing must be lawful, fair, and transparent to the data subject.
    \item Purpose limitation — You must process data for the legitimate purposes specified explicitly to the data subject when you collected it.
    \item Data minimization — You should collect and process only as much data as absolutely necessary for the purposes specified.
    \item Accuracy — You must keep personal data accurate and up to date.
    \item Storage limitation — You may only store personally identifying data for as long as necessary for the specified purpose.
    \item Integrity and confidentiality — Processing must be done in such a way as to ensure appropriate security, integrity, and confidentiality (e.g. by using encryption).
    \item Accountability — The data controller is responsible for being able to demonstrate GDPR compliance with all of these principles.
\end{itemize}

The legal response of BIPA and GDPR can be in part traced to the rapid proliferation of biometric images (especially face) on social media websites such as Facebook, Twitter and Instagram, and their use in training biometric recognition systems without the informed consent of subjects. For example, a face recognition startup, Clearview AI, is currently facing litigation for allegedly amassing a dataset of about $3$ billion face images~\cite{clearview} from various social media sites without subjects' permission. This lack of consent and transparency has led to some cities wanting to curb facial recognition technology\footnote{\url{https://www.wcjb.com/2021/05/05/states-push-back-against-use-of-facial-recognition-by-police/}}~\footnote{\url{https://www.usatoday.com/story/tech/2019/12/17/face-recognition-ban-some-cities-states-and-lawmakers-push-one/2680483001/}}. In addition, publicly available biometric datasets that were collected without consent are now being retracted~\cite{MSceleb_retracted,duke_mtmc, brainwash, watson1992nist, watson1993nist}. To make matters worse, there has been work to show that even generative adversarial networks can leak private information about the dataset on which they were trained~\cite{tinsley2021face}.

In the computer vision community, research has been conducted to alleviate these concerns. In particular, a number of studies have explored using homomorphic encryption to perform inference or classification on encrypted data with encrypted model parameters~\cite{nalini1, nalini2, nalini3, gilad2016cryptonets, brutzkus2019low, yonetani2017privacy} (see Figure~\ref{fig:overview_privacy}). While these approaches are quite promising as they offer the data/model parameters a high level of security and consequently privacy, they require significant computational burden which limits the size of models in practice. Basically, as with our previous discussion on trade offs of speed, memory, accuracy and security when using fully homomorphic encryption for protecting biometric templates, the same issue applies towards its use in protecting data and model parameters. 

An alternative method that has been explored to impart privacy is to train biometric systems in a decentralized manner (\ie federated learning). In particular, multiple participating clients jointly learn a biometric recognition system without ever sharing their training data with each other. For example the study in~\cite{shao2020federated} used federated learning for training a face PAD algorithm. Likewise, Aggarwal \etal~\cite{fedface} propose to use federated learning to collaboratively learn a global face recognition system, training from face images on multiple clients (mobile devices) in a privacy preserving manner. Only local updates from each mobile device are shared to the server where they are aggregated and used to optimize the global objective function, while the training face images on each mobile device are kept private. Their proposed framework is able to enhance the performance of a pretrained face recognition system namely, CosFace~\cite{cosface}, from a TAR of $81.43\% \xrightarrow{} 83.79\%$ on IJB-A dataset~\cite{ijba} at  $0.1\%$ FAR and accuracy from $99.15\% \xrightarrow{} 99.28\%$ on LFW dataset~\cite{lfw} using the face images available on $1,000$ mobile devices in a federated setup. A limitation of this approach is that the performance of the AFR when using decentralized training is inferior to a centralized training schema, \ie recognition performance is traded off for privacy. Therefore, future work is required to reduce this trade off.

We encourage further exploration of encrypted inference methods and also decentralized training methods to continue to enhance the privacy of biometric recognition systems. Given the legal ramifications and public awareness surrounding this topic, further improvements in this area of research are paramount towards achieving trustworthy biometric recognition systems.

%% file: ch_conclusion.tex
\section{Conclusion}

Accurate and reliable automatic person identification is becoming a necessity in a host of applications including national ID cards, border crossings, access control, payments, \etc. Biometric recognition stands as perhaps the most well equipped technology to meet this need. Indeed, biometric recognition systems have now matured to the point at which they can surpass human recognition performance or accuracy under certain conditions. However, many unsolved problems remain prior to acceptance of biometric recognition systems as \textit{trustworthy}. In this paper, we have highlighted five major areas of research that must be further worked on in order to establish trustworthiness in biometrics: 1) Performance Robustness and Scalability, 2) Security, 3) Explainability and Interpretability, 4) Biasness and Fairness, and 5) Privacy. In each of these areas, we have provided a problem definition, explained the importance of the problem, cited existing work on each respective topic, and concluded with suggestions for further research. By better addressing each of these major areas, biometric recognition systems can be made not only accurate, but also trustworthy. This benefits the researchers behind the recognition systems, the general public using the systems, and the policy makers regulating the systems. 

One practical potential avenue to encourage more trustworthy biometric recognition systems is a ``Grand Challenge on Trustworthy Biometrics". Perhaps such a challenge, hosted by a government agency, say NIST, could evaluate the biometric recognition systems on each of the 5 categories listed above. Systems that met certain quantitative thresholds for the 5 categories above could be certified as ``trustworthy". In this manner, end-users would know not only how accurate the system is, but also how ``trustworthy" it is.

%% file: main.bbl
\begin{thebibliography}{100}

\bibitem{surveillance}
{The Guardian}, ``{Commuters walk by surveillance cameras at a walkway between
  two subway stations in Beijing}.''
  \url{https://www.theguardian.com/cities/2019/dec/02/big-brother-is-watching-chinese-city-with-26m-cameras-is-worlds-most-heavily-surveilled},
  2019.
\newblock [Online; accessed 11-April-2021].

\bibitem{ecommerce}
{PC guide}, ``{Amazon Announces New Contactless Payment System Using Just Your
  Hand}.''
  \url{https://www.pcguide.com/news/amazon-announces-new-contactless-payment-system-using-just-your-hand/},
  2020.
\newblock [Online; accessed 11-April-2021].

\bibitem{boarding}
{CNN}, ``{How facial recognition is taking over airports}.''
  \url{https://www.cnn.com/travel/article/airports-facial-recognition/index.html},
  2019.
\newblock [Online; accessed 17-April-2021].

\bibitem{matchoncard}
{Shutterstock}, ``{Credit card with a fingerprint sensor}.''
  \url{https://www.shutterstock.com/image-photo/credit-card-fingerprint-sensor-purchase-biometric-1020927472},
  2020.
\newblock [Online; accessed 11-April-2021].

\bibitem{atm}
{el Boletin}, ``{CaixaBank, the first bank in the world to use facial
  recognition in its ATMs}.''
  \url{https://www.elboletin.com/caixabank-primer-banco-del-mundo-que-utiliza-el-reconocimiento-facial-en-sus-cajeros/},
  2019.
\newblock [Online; accessed 11-April-2021].

\bibitem{smartphone}
{Techspot}, ``{iPhone X users are inadvertently training Face ID to recognize
  siblings}.''
  \url{https://www.techspot.com/news/71741-iphone-x-users-inadvertently-training-face-recognize-siblings.html},
  2017.
\newblock [Online; accessed 11-April-2021].

\bibitem{bordercontrol}
{Gerald Nino}, ``{U.S. Customs and Border Protection}.''
  \url{https://commons.wikimedia.org/wiki/File:US-VISIT_(CBP).jpg}, 2007.
\newblock [Online; accessed 11-April-2021].

\bibitem{socialbenefits}
{SRBPost}, ``{Ration Card Aadhaar Card Link Online}.''
  \url{http://www.srbpost.com/sarkari-yojana/ration-card-aadhaar-card-link-online/},
  2020.
\newblock [Online; accessed 11-April-2021].

\bibitem{coalminers}
{China Daily}, ``{Global vision drives iris-recognition technology}.''
  \url{https://www.chinadaily.com.cn/a/201809/14/WS5b9b0fbca31033b4f4655fe8.html},
  2018.
\newblock [Online; accessed 17-April-2021].

\bibitem{vehicular}
{L'argus}, ``{Fingerprint reader and facial recognition camera}.''
  \url{https://www.largus.fr/actualite-automobile/ces-2017-biometrie-et-conduite-autonome-au-menu-pour-continental-8325393.html},
  2017.
\newblock [Online; accessed 11-April-2021].

\bibitem{amazon_one_news}
{The Verge}, ``{Amazon’s palm reading starts at the grocery store, but it
  could be so much bigger}.''
  \url{https://www.theverge.com/2020/10/1/21496673/amazon-one-palm-reading-vein-recognition-payments-identity-verification},
  2020.
\newblock [Online; accessed 17-April-2021].

\bibitem{us_visit}
{Department of Homeland Security}, ``{US-VISIT Face Sheet}.''
  \url{https://www.dhs.gov/xlibrary/assets/usvisit/usvisit_edu_10-fingerprint_collection_fact_sheet.pdf},
  2009.
\newblock [Online; accessed 17-April-2021].

\bibitem{jain2011introduction}
A.~K. Jain, A.~A. Ross, and K.~Nandakumar, {\em Introduction to biometrics}.
\newblock Springer Science \& Business Media, 2011.

\bibitem{habitual_criminals}
{Victorian Crime \& Punishment}, ``{Habitual Criminals}.''
  \url{http://vcp.e2bn.org/justice/page11571-habitual-criminals.html}, 2006.
\newblock [Online; accessed 25-April-2021].

\bibitem{home_office_quote}
A.~Barrett and C.~Harrison, {\em Crime and punishment in England: A
  sourcebook}.
\newblock Routledge, 2005.

\bibitem{ancient_rome}
T.~R. Martin, {\em Ancient Rome: From Romulus to Justinian}.
\newblock Yale University Press, 2012.

\bibitem{social_engineering}
{IT Pro}, ``{Fraudsters use AI voice manipulation to steal £200,000}.''
  \url{https://www.itpro.com/social-engineering/34308/fraudsters-use-ai-voice-manipulation-to-steal-200000},
  2019.
\newblock [Online; accessed 21-April-2021].

\bibitem{keylogging}
{The Quint}, ``{Hackers Can Detect What You’re Typing By Listening To You
  Type}.''
  \url{https://www.thequint.com/tech-and-auto/tech-news/hackers-can-know-your-password-just-by-listening-to-your-typing},
  2019.
\newblock [Online; accessed 21-April-2021].

\bibitem{msn}
{Microsoft News}, ``{WhatsApp is leaking mobile numbers of users in
  plaintext}.''
  \url{https://www.msn.com/en-in/money/news/whatsapp-is-leaking-mobile-numbers-of-users-in-plaintext-claims-an-independent-cybersecurity-researcher/ar-BB15e8Ak?li=AAggbRN},
  2020.
\newblock [Online; accessed 21-April-2021].

\bibitem{smudge_attacks}
A.~J. Aviv, K.~L. Gibson, E.~Mossop, M.~Blaze, and J.~M. Smith, ``Smudge
  attacks on smartphone touch screens.,'' {\em Woot}, vol.~10, pp.~1--7, 2010.

\bibitem{fraud}
{World Health Organization}, ``{Prevention not cure in tackling health-care
  fraud}.'' \url{https://www.who.int/bulletin/volumes/89/12/11-021211/en/},
  2011.
\newblock [Online; accessed 11-April-2021].

\bibitem{fraud2}
{World Food Programme}, ``{WFP demands action after uncovering misuse of food
  relief intended for hungry people in Yemen}.''
  \url{https://www1.wfp.org/news/wfp-demands-action-after-uncovering-misuse-food-relief-intended-hungry-people-yemen},
  2018.
\newblock [Online; accessed 11-April-2021].

\bibitem{fraud3}
{World Food Programme Insight}, ``{These changes show that WFP loves us}.''
  \url{https://insight.wfp.org/these-changes-show-that-wfp-loves-us-247f0c1ebcf},
  2018.
\newblock [Online; accessed 13-April-2021].

\bibitem{dhiren}
{Evening Standard}, ``{Bin Laden's general was given NINE British passports}.''
  \url{https://www.standard.co.uk/hp/front/bin-laden-s-general-was-given-nine-british-passports-7170522.html},
  2012.
\newblock [Online; accessed 16-April-2021].

\bibitem{stevenson2010oxford}
A.~Stevenson, {\em Oxford dictionary of English}.
\newblock Oxford University Press, USA, 2010.

\bibitem{trauring1963automatic}
M.~Trauring, ``Automatic comparison of finger-ridge patterns,'' {\em Nature},
  vol.~197, no.~4871, pp.~938--940, 1963.

\bibitem{iafis}
{Federal Bureau of Investigation}, ``{Integrated Automated Fingerprint
  Identification System}.''
  \url{https://www.fbi.gov/services/information-management/foipa/privacy-impact-assessments/iafis},
  2021.
\newblock [Online; accessed 27-April-2021].

\bibitem{deep_learning_biometrics1}
K.~Sundararajan and D.~L. Woodard, ``Deep learning for biometrics: A survey,''
  {\em ACM Computing Surveys}, vol.~51, no.~3, pp.~1--34, 2018.

\bibitem{deep_learning_biometrics2}
B.~Bhanu, A.~Kumar, {\em et~al.}, {\em Deep learning for biometrics}.
\newblock Springer, 2017.

\bibitem{deep_learning_biometrics3}
M.~Vatsa, R.~Singh, and A.~Majumdar, {\em Deep Learning in Biometrics}.
\newblock CRC Press, 2018.

\bibitem{lu2015surpassing}
C.~Lu and X.~Tang, ``Surpassing human-level face verification performance on
  lfw with gaussianface,'' in {\em AAAI}, vol.~29, 2015.

\bibitem{nist_fpvte}
C.~I. Watson, G.~P. Fiumara, E.~Tabassi, S.~L. Cheng, P.~A. Flanagan, and W.~J.
  Salamon, ``Fingerprint vendor technology evaluation,'' {\em NIST}, 2015.

\bibitem{grother2018ongoing}
P.~J. Grother, M.~L. Ngan, and K.~K. Hanaoka, ``Ongoing face recognition vendor
  test (frvt) part 2: Identification,'' {\em NIST}, 2018.

\bibitem{quinn2019irex}
G.~W. Quinn and J.~R. Matey, {\em IREX 10: Ongoing Evaluation of Iris
  Recognition Concept, Evaluation Plan, and API Overview}.
\newblock NIST, 2019.

\bibitem{marasco2014survey}
E.~Marasco and A.~Ross, ``A survey on antispoofing schemes for fingerprint
  recognition systems,'' {\em ACM Computing Surveys (CSUR)}, vol.~47, no.~2,
  pp.~1--36, 2014.

\bibitem{singh2020survey}
J.~M. Singh, A.~Madhun, G.~Li, and R.~Ramachandra, ``A survey on unknown
  presentation attack detection for fingerprint,'' {\em arXiv preprint
  arXiv:2005.08337}, 2020.

\bibitem{jain2008biometric}
A.~K. Jain, K.~Nandakumar, and A.~Nagar, ``Biometric template security,'' {\em
  EURASIP Journal on advances in signal processing}, vol.~2008, pp.~1--17,
  2008.

\bibitem{jain_50}
A.~K. Jain, K.~Nandakumar, and A.~Ross, ``50 years of biometric research:
  Accomplishments, challenges, and opportunities,'' {\em Pattern recognition
  letters}, vol.~79, pp.~80--105, 2016.

\bibitem{ross2019some}
A.~Ross, S.~Banerjee, C.~Chen, A.~Chowdhury, V.~Mirjalili, R.~Sharma,
  T.~Swearingen, and S.~Yadav, ``Some research problems in biometrics: The
  future beckons,'' in {\em ICB}, pp.~1--8, IEEE, 2019.

\bibitem{jain2015bridging}
A.~K. Jain and A.~Ross, ``Bridging the gap: from biometrics to forensics,''
  {\em Philosophical Transactions of the Royal Society B: Biological Sciences},
  vol.~370, no.~1674, p.~20140254, 2015.

\bibitem{vakhshiteh2020adversarial}
F.~Vakhshiteh, A.~Nickabadi, and R.~Ramachandra, ``Adversarial attacks against
  face recognition: A comprehensive study,'' {\em arXiv preprint
  arXiv:2007.11709}, 2020.

\bibitem{drozdowski2020demographic}
P.~Drozdowski, C.~Rathgeb, A.~Dantcheva, N.~Damer, and C.~Busch, ``Demographic
  bias in biometrics: A survey on an emerging challenge,'' {\em IEEE
  Transactions on Technology and Society}, vol.~1, no.~2, pp.~89--103, 2020.

\bibitem{boyd2020iris}
A.~Boyd, Z.~Fang, A.~Czajka, and K.~W. Bowyer, ``Iris presentation attack
  detection: Where are we now?,'' {\em Pattern Recognition Letters}, vol.~138,
  pp.~483--489, 2020.

\bibitem{fvc}
B.~Dorizzi, R.~Cappelli, M.~Ferrara, D.~Maio, D.~Maltoni, N.~Houmani,
  S.~Garcia-Salicetti, and A.~Mayoue, ``Fingerprint and on-line signature
  verification competition,'' in {\em IEEE ICB}, pp.~725--732, 2009.

\bibitem{lfw}
G.~B. Huang, M.~Ramesh, T.~Berg, and E.~Learned-Miller, ``Labeled faces in the
  wild: A database for studying face recognition in unconstrained
  environments,'' Tech. Rep. 07-49, University of Massachusetts, Amherst,
  October 2007.

\bibitem{wang2019toward}
K.~Wang and A.~Kumar, ``Toward more accurate iris recognition using dilated
  residual features,'' {\em IEEE TIFS}, vol.~14, no.~12, pp.~3233--3245, 2019.

\bibitem{market_share}
{Mordor Intelligence}, ``Consumer biometrics market-growth.''
  \url{https://www.mordorintelligence.com/industry-reports/consumer-biometrics-market},
  2021.
\newblock [Online; accessed 13-May-2021].

\bibitem{chugh2017benchmarking}
T.~Chugh, S.~S. Arora, A.~K. Jain, and N.~G. Paulter, ``Benchmarking
  fingerprint minutiae extractors,'' in {\em IEEE BIOSIG}, pp.~1--8, 2017.

\bibitem{grosz2020white}
S.~A. Grosz, J.~J. Engelsma, and A.~K. Jain, ``White-box evaluation of
  fingerprint recognition systems,'' {\em arXiv preprint arXiv:2008.00128},
  2020.

\bibitem{bowyer2016handbook}
K.~W. Bowyer and M.~J. Burge, {\em Handbook of iris recognition}.
\newblock Springer, 2016.

\bibitem{grother2019frvt}
P.~Grother, M.~Ngan, and K.~Hanaoka, ``Face recognition vendor test ({FRVT})
  part 3: Demographic effects,'' in {\em NIST}, 2019.

\bibitem{engelsma2021infant}
J.~J. Engelsma, D.~Deb, K.~Cao, A.~Bhatnagar, P.~S. Sudhish, and A.~K. Jain,
  ``Infant-id: Fingerprints for global good,'' {\em IEEE PAMI}, 2021.

\bibitem{si2015detection}
X.~Si, J.~Feng, J.~Zhou, and Y.~Luo, ``Detection and rectification of distorted
  fingerprints,'' {\em IEEE PAMI}, vol.~37, no.~3, pp.~555--568, 2015.

\bibitem{cao2019end}
K.~Cao, D.-L. Nguyen, C.~Tymoszek, and A.~K. Jain, ``End-to-end latent
  fingerprint search,'' {\em IEEE TIFS}, vol.~15, pp.~880--894, 2019.

\bibitem{engelsma2019learning}
J.~J. Engelsma, K.~Cao, and A.~K. Jain, ``Learning a fixed-length fingerprint
  representation,'' {\em IEEE PAMI}, 2019.

\bibitem{tran2017disentangled}
L.~Tran, X.~Yin, and X.~Liu, ``Disentangled representation learning gan for
  pose-invariant face recognition,'' in {\em CVPR}, pp.~1415--1424, 2017.

\bibitem{best2017longitudinal}
L.~Best-Rowden and A.~K. Jain, ``Longitudinal study of automatic face
  recognition,'' {\em IEEE PAMI}, vol.~40, no.~1, pp.~148--162, 2017.

\bibitem{johnson2018longitudinal}
M.~Johnson, D.~Yambay, D.~Rissacher, L.~Holsopple, and S.~Schuckers, ``A
  longitudinal study of iris recognition in children,'' in {\em IEEE ISBA},
  pp.~1--7, 2018.

\bibitem{yoon2015longitudinal}
S.~Yoon and A.~K. Jain, ``Longitudinal study of fingerprint recognition,'' {\em
  Proceedings of the National Academy of Sciences}, vol.~112, no.~28,
  pp.~8555--8560, 2015.

\bibitem{irex_vi}
P.~J. Grother, J.~R. Matey, E.~Tabassi, G.~W. Quinn, and M.~Chumakov, ``Irex
  vi-temporal stability of iris recognition accuracy,'' {\em NIST}, 2013.

\bibitem{anatomy_face}
S.~R. Coleman and R.~Grover, ``{The Anatomy of the Aging Face: Volume Loss and
  Changes in 3-Dimensional Topography},'' {\em Aesthetic Surgery Journal},
  vol.~26, pp.~S4--S9, 2006.

\bibitem{facialstructure}
N.~Ramanathan and R.~Chellappa, ``Modeling age progression in young faces,'' in
  {\em CVPR}, 2006.

\bibitem{klare2012face}
B.~F. Klare, M.~J. Burge, J.~C. Klontz, W.~V. Bruegge, Richard, and A.~K. Jain,
  ``Face recognition performance: Role of demographic information,'' {\em IEEE
  TIFS}, vol.~7, no.~6, pp.~1789--1801, 2012.

\bibitem{deb2017face}
D.~Deb, L.~Best-Rowden, and A.~K. Jain, ``Face recognition performance under
  aging,'' in {\em CVPRW}, 2017.

\bibitem{deb2018longitudinal}
D.~Deb, N.~Nain, and A.~K. Jain, ``Longitudinal study of child face
  recognition,'' in {\em IEEE ICB}, pp.~225--232, 2018.

\bibitem{deb2020child}
D.~Deb, D.~Aggarwal, and A.~K. Jain, ``Identifying missing children: Face age
  progression via deep feature aging,'' {\em IEEE ICPR}, 2020.

\bibitem{cosface}
H.~Wang, Y.~Wang, Z.~Zhou, X.~Ji, D.~Gong, J.~Zhou, Z.~Li, and W.~Liu,
  ``Cosface: Large margin cosine loss for deep face recognition,'' in {\em
  CVPR}, 2018.

\bibitem{arcface}
J.~Deng, J.~Guo, N.~Xue, and S.~Zafeiriou, ``Arcface: Additive angular margin
  loss for deep face recognition,'' in {\em CVPR}, pp.~4690--4699, 2019.

\bibitem{MsCeleb}
Y.~Guo, L.~Zhang, Y.~Hu, X.~He, and J.~Gao, ``Ms-celeb-1m: A dataset and
  benchmark for large-scale face recognition,'' in {\em ECCV}, pp.~87--102,
  2016.

\bibitem{yin2019feature}
X.~Yin, X.~Yu, K.~Sohn, X.~Liu, and M.~Chandraker, ``Feature transfer learning
  for face recognition with under-represented data,'' in {\em CVPR},
  pp.~5704--5713, 2019.

\bibitem{wan2018cost}
J.~Wan and Y.~Wang, ``Cost-sensitive label propagation for semi-supervised face
  recognition,'' {\em IEEE TIFS}, vol.~14, no.~7, pp.~1729--1743, 2018.

\bibitem{yang2020learning}
L.~Yang, D.~Chen, X.~Zhan, R.~Zhao, C.~C. Loy, and D.~Lin, ``Learning to
  cluster faces via confidence and connectivity estimation,'' in {\em CVPR},
  pp.~13369--13378, 2020.

\bibitem{guo2020density}
S.~Guo, J.~Xu, D.~Chen, C.~Zhang, X.~Wang, and R.~Zhao, ``Density-aware feature
  embedding for face clustering,'' in {\em CVPR}, pp.~6698--6706, 2020.

\bibitem{roychowdhury2020improving}
A.~RoyChowdhury, X.~Yu, K.~Sohn, E.~Learned-Miller, and M.~Chandraker,
  ``Improving face recognition by clustering unlabeled faces in the wild,'' in
  {\em ECCV}, pp.~119--136, 2020.

\bibitem{zhang2020neighborhood}
Q.~Zhang, Z.~Lei, and S.~Z. Li, ``Neighborhood-aware attention network for
  semi-supervised face recognition,'' in {\em IEEE IJCNN}, pp.~1--8, 2020.

\bibitem{shi2021boosting}
Y.~Shi and A.~K. Jain, ``Boosting unconstrained face recognition with auxiliary
  unlabeled data,'' {\em CVPR Workshops}, 2021.

\bibitem{aadhar}
R.~S. Sharma, {\em {THE MAKING OF AADHAAR: World’s Largest Identity
  Platform}}.
\newblock Rupa Publications India, 2020.

\bibitem{fbi_ngi}
{Federal Bureau of Investigation}, ``{FBI Announces Contract Award for Next
  Generation Identification System}.''
  \url{https://archives.fbi.gov/archives/news/pressrel/press-releases/fbi-announces-contract-award-for-next-generation-identification-system},
  2008.
\newblock [Online; accessed 27-April-2021].

\bibitem{dhs}
{U.S. Department of Homeland Security}, ``{DHS/CBP/PIA-014 Centralized Area
  Video Surveillance System}.''
  \url{https://www.dhs.gov/publication/centralized-area-video-surveillance-system},
  2013.
\newblock [Online; accessed 12-May-2021].

\bibitem{daugman2003importance}
J.~Daugman, ``The importance of being random: statistical principles of iris
  recognition,'' {\em Pattern recognition}, vol.~36, no.~2, pp.~279--291, 2003.

\bibitem{wang2016face}
D.~Wang, C.~Otto, and A.~K. Jain, ``Face search at scale,'' {\em IEEE PAMI},
  vol.~39, no.~6, pp.~1122--1136, 2016.

\bibitem{otto2017clustering}
C.~Otto, D.~Wang, and A.~K. Jain, ``Clustering millions of faces by identity,''
  {\em IEEE PAMI}, vol.~40, no.~2, pp.~289--303, 2017.

\bibitem{mistry2019fingerprint}
V.~Mistry, J.~J. Engelsma, and A.~K. Jain, ``Fingerprint synthesis: Search with
  100 million prints,'' in {\em IEEE IJCB}, pp.~1--10, 2019.

\bibitem{yoon2012altered}
S.~Yoon, J.~Feng, and A.~K. Jain, ``Altered fingerprints: Analysis and
  detection,'' {\em IEEE PAMI}, vol.~34, no.~3, pp.~451--464, 2012.

\bibitem{arora2016design}
S.~S. Arora, K.~Cao, A.~K. Jain, and N.~G. Paulter, ``Design and fabrication of
  3d fingerprint targets,'' {\em IEEE TIFS}, vol.~11, no.~10, pp.~2284--2297,
  2016.

\bibitem{engelsma2018universal}
J.~J. Engelsma, S.~S. Arora, A.~K. Jain, and N.~G. Paulter, ``Universal 3d
  wearable fingerprint targets: advancing fingerprint reader evaluations,''
  {\em IEEE TIFS}, vol.~13, no.~6, pp.~1564--1578, 2018.

\bibitem{schultz2021three}
C.~W. Schultz, M.~Fawzy, F.~Nasirpouri, K.~L. Kavanagh, and H.-Z. Yu,
  ``Three-dimensional conductive fingerprint phantoms made of ethylene-vinyl
  acetate/graphene nanocomposite for evaluating smartphone scanners,'' {\em ACS
  Applied Electronic Materials}, 2021.

\bibitem{cao2016hacking}
K.~Cao and A.~K. Jain, ``Hacking mobile phones using 2d printed fingerprints,''
  {\em Dept. Comput. Sci. Eng., Michigan State Univ., East Lansing, MI, USA,
  Tech. Rep. MSU-CSE-16-2}, 2016.

\bibitem{schuckers2002spoofing}
S.~A. Schuckers, ``Spoofing and anti-spoofing measures,'' {\em Information
  Security technical report}, vol.~7, no.~4, pp.~56--62, 2002.

\bibitem{liu2019deep}
Y.~Liu, J.~Stehouwer, A.~Jourabloo, and X.~Liu, ``Deep tree learning for
  zero-shot face anti-spoofing,'' in {\em CVPR}, pp.~4680--4689, 2019.

\bibitem{chugh2019fingerprint}
T.~Chugh and A.~K. Jain, ``Fingerprint presentation attack detection:
  Generalization and efficiency,'' in {\em IEEE ICB}, pp.~1--8, 2019.

\bibitem{hoffman2018convolutional}
S.~Hoffman, R.~Sharma, and A.~Ross, ``Convolutional neural networks for iris
  presentation attack detection: Toward cross-dataset and cross-sensor
  generalization,'' in {\em CVPRW}, pp.~1620--1628, 2018.

\bibitem{van2000biometrical}
T.~Van~der Putte and J.~Keuning, ``Biometrical fingerprint recognition: don’t
  get your fingers burned,'' in {\em Smart Card Research and Advanced
  Applications}, pp.~289--303, Springer, 2000.

\bibitem{matsumoto2002impact}
T.~Matsumoto, H.~Matsumoto, K.~Yamada, and S.~Hoshino, ``Impact of artificial"
  gummy" fingers on fingerprint systems,'' in {\em Optical Security and
  Counterfeit Deterrence Techniques IV}, vol.~4677, pp.~275--289, 2002.

\bibitem{orru2019livdet}
G.~Orr{\`u}, R.~Casula, P.~Tuveri, C.~Bazzoni, G.~Dessalvi, M.~Micheletto,
  L.~Ghiani, and G.~L. Marcialis, ``Livdet in action-fingerprint liveness
  detection competition 2019,'' in {\em IEEE ICB}, pp.~1--6, 2019.

\bibitem{das2020iris}
P.~Das, J.~Mcfiratht, Z.~Fang, A.~Boyd, G.~Jang, A.~Mohammadi, S.~Purnapatra,
  D.~Yambay, S.~Marcel, M.~Trokielewicz, {\em et~al.}, ``Iris liveness
  detection competition (livdet-iris)-the 2020 edition,'' in {\em IEEE IJCB},
  pp.~1--9, 2020.

\bibitem{zhang2021celeba}
Y.~Zhang, Z.~Yin, J.~Shao, Z.~Liu, S.~Yang, Y.~Xiong, W.~Xia, Y.~Xu, M.~Luo,
  J.~Liu, {\em et~al.}, ``Celeba-spoof challenge 2020 on face anti-spoofing:
  Methods and results,'' {\em arXiv preprint arXiv:2102.12642}, 2021.

\bibitem{ramachandra2017presentation}
R.~Ramachandra and C.~Busch, ``Presentation attack detection methods for face
  recognition systems: A comprehensive survey,'' {\em ACM Computing Surveys
  (CSUR)}, vol.~50, no.~1, pp.~1--37, 2017.

\bibitem{galbally2014biometric}
J.~Galbally, S.~Marcel, and J.~Fierrez, ``Biometric antispoofing methods: A
  survey in face recognition,'' {\em IEEE Access}, vol.~2, pp.~1530--1552,
  2014.

\bibitem{marcel2019handbook}
S.~Marcel, M.~S. Nixon, J.~Fierrez, and N.~Evans, {\em Handbook of biometric
  anti-spoofing: Presentation attack detection}.
\newblock Springer, 2019.

\bibitem{baldisserra2006fake}
D.~Baldisserra, A.~Franco, D.~Maio, and D.~Maltoni, ``Fake fingerprint
  detection by odor analysis,'' in {\em ICB}, pp.~265--272, 2006.

\bibitem{nixon2008spoof}
K.~A. Nixon, V.~Aimale, and R.~K. Rowe, ``Spoof detection schemes,'' in {\em
  Handbook of biometrics}, pp.~403--423, Springer, 2008.

\bibitem{tolosana2018towards}
R.~Tolosana, M.~Gomez-Barrero, J.~Kolberg, A.~Morales, C.~Busch, and
  J.~Ortega-Garcia, ``Towards fingerprint presentation attack detection based
  on convolutional neural networks and short wave infrared imaging,'' in {\em
  IEEE BIOSIG}, pp.~1--5, 2018.

\bibitem{keilbach2018fingerprint}
P.~Keilbach, J.~Kolberg, M.~Gomez-Barrero, C.~Busch, and H.~Langweg,
  ``Fingerprint presentation attack detection using laser speckle contrast
  imaging,'' in {\em IEEE BIOSIG}, pp.~1--6, 2018.

\bibitem{hussein2018fingerprint}
M.~E. Hussein, L.~Spinoulas, F.~Xiong, and W.~Abd-Almageed, ``Fingerprint
  presentation attack detection using a novel multi-spectral capture device and
  patch-based convolutional neural networks,'' in {\em IEEE International
  Workshop on Information Forensics and Security}, pp.~1--8, 2018.

\bibitem{engelsma2018raspireader}
J.~J. Engelsma, K.~Cao, and A.~K. Jain, ``Raspireader: Open source fingerprint
  reader,'' {\em IEEE PAMI}, vol.~41, no.~10, pp.~2511--2524, 2018.

\bibitem{wang2013face}
T.~Wang, J.~Yang, Z.~Lei, S.~Liao, and S.~Z. Li, ``Face liveness detection
  using 3d structure recovered from a single camera,'' in {\em IEEE ICB},
  pp.~1--6, 2013.

\bibitem{wang2017robust}
Y.~Wang, F.~Nian, T.~Li, Z.~Meng, and K.~Wang, ``Robust face anti-spoofing with
  depth information,'' {\em Journal of Visual Communication and Image
  Representation}, vol.~49, pp.~332--337, 2017.

\bibitem{conotter2014physiologically}
V.~Conotter, E.~Bodnari, G.~Boato, and H.~Farid, ``Physiologically-based
  detection of computer generated faces in video,'' in {\em IEEE International
  Conference on Image Processing}, pp.~248--252, 2014.

\bibitem{zhang2011face}
Z.~Zhang, D.~Yi, Z.~Lei, and S.~Z. Li, ``Face liveness detection by learning
  multispectral reflectance distributions,'' in {\em IEEE Face and Gesture},
  pp.~436--441, 2011.

\bibitem{heusch2020deep}
G.~Heusch, A.~George, D.~Geissb{\"u}hler, Z.~Mostaani, and S.~Marcel, ``Deep
  models and shortwave infrared information to detect face presentation
  attacks,'' {\em IEEE T-BIOM}, vol.~2, no.~4, pp.~399--409, 2020.

\bibitem{fang2020open}
Z.~Fang and A.~Czajka, ``Open source iris recognition hardware and software
  with presentation attack detection,'' in {\em IEEE IJCB}, pp.~1--8, 2020.

\bibitem{george2019deep}
A.~George and S.~Marcel, ``Deep pixel-wise binary supervision for face
  presentation attack detection,'' in {\em IEEE ICB}, 2019.

\bibitem{menotti2015deep}
D.~Menotti, G.~Chiachia, A.~Pinto, W.~R. Schwartz, H.~Pedrini, A.~X. Falcao,
  and A.~Rocha, ``Deep representations for iris, face, and fingerprint spoofing
  detection,'' {\em IEEE TIFS}, vol.~10, no.~4, pp.~864--879, 2015.

\bibitem{yadav2020relativistic}
S.~Yadav, C.~Chen, and A.~Ross, ``Relativistic discriminator: A one-class
  classifier for generalized iris presentation attack detection,'' in {\em
  WACV}, 2020.

\bibitem{tolosana2019biometric}
R.~Tolosana, M.~Gomez-Barrero, C.~Busch, and J.~Ortega-Garcia, ``Biometric
  presentation attack detection: Beyond the visible spectrum,'' {\em IEEE
  TIFS}, vol.~15, pp.~1261--1275, 2019.

\bibitem{yadav2021cit}
S.~Yadav and A.~Ross, ``Cit-gan: Cyclic image translation generative
  adversarial network with application in iris presentation attack detection,''
  in {\em WACV}, 2021.

\bibitem{hoffman2019iris+}
S.~Hoffman, R.~Sharma, and A.~Ross, ``Iris+ ocular: Generalized iris
  presentation attack detection using multiple convolutional neural networks,''
  in {\em IEEE ICB}, 2019.

\bibitem{sharma2021viability}
R.~Sharma and A.~Ross, ``Viability of optical coherence tomography for iris
  presentation attack detection,'' in {\em IEEE ICPR}, 2021.

\bibitem{czajka2018presentation}
A.~Czajka and K.~W. Bowyer, ``Presentation attack detection for iris
  recognition: An assessment of the state-of-the-art,'' {\em ACM Computing
  Surveys}, vol.~51, no.~4, pp.~1--35, 2018.

\bibitem{morales2019introduction}
A.~Morales, J.~Fierrez, J.~Galbally, and M.~Gomez-Barrero, ``Introduction to
  iris presentation attack detection,'' in {\em Handbook of Biometric
  Anti-Spoofing}, pp.~135--150, Springer, 2019.

\bibitem{ferreira2019adversarial}
P.~M. Ferreira, A.~F. Sequeira, D.~Pernes, A.~Rebelo, and J.~S. Cardoso,
  ``Adversarial learning for a robust iris presentation attack detection method
  against unseen attack presentations,'' in {\em IEEE BIOSIG}, pp.~1--7, 2019.

\bibitem{engelsma2019generalizing}
J.~J. Engelsma and A.~K. Jain, ``Generalizing fingerprint spoof detector:
  Learning a one-class classifier,'' in {\em IEEE ICB}, pp.~1--8, 2019.

\bibitem{liu2018learning}
Y.~Liu, A.~Jourabloo, and X.~Liu, ``Learning deep models for face
  anti-spoofing: Binary or auxiliary supervision,'' in {\em CVPR},
  pp.~389--398, 2018.

\bibitem{deb2020look}
D.~Deb and A.~K. Jain, ``Look locally infer globally: A generalizable face
  anti-spoofing approach,'' {\em IEEE TIFS}, vol.~16, pp.~1143--1157, 2020.

\bibitem{jaiswal2019ropad}
A.~Jaiswal, S.~Xia, I.~Masi, and W.~AbdAlmageed, ``Ropad: Robust presentation
  attack detection through unsupervised adversarial invariance,'' in {\em IEEE
  ICB}, pp.~1--8, 2019.

\bibitem{gajawada2019universal}
R.~Gajawada, A.~Popli, T.~Chugh, A.~Namboodiri, and A.~K. Jain, ``Universal
  material translator: Towards spoof fingerprint generalization,'' in {\em IEEE
  ICB}, pp.~1--8, 2019.

\bibitem{kolberg2021anomaly}
J.~Kolberg, M.~Grimmer, M.~Gomez-Barrero, and C.~Busch, ``Anomaly detection
  with convolutional autoencoders for fingerprint presentation attack
  detection,'' {\em IEEE T-BIOM}, vol.~3, no.~2, pp.~190--202, 2021.

\bibitem{george2020learning}
A.~George and S.~Marcel, ``Learning one class representations for face
  presentation attack detection using multi-channel convolutional neural
  networks,'' {\em IEEE TIFS}, vol.~16, pp.~361--375, 2020.

\bibitem{kolberg2020generalisation}
J.~Kolberg, M.~Gomez-Barrero, and C.~Busch, ``On the generalisation
  capabilities of fingerprint presentation attack detection methods in the
  short wave infrared domain,'' {\em arXiv preprint arXiv:2010.09566}, 2020.

\bibitem{george2020effectiveness}
A.~George and S.~Marcel, ``On the effectiveness of vision transformers for
  zero-shot face anti-spoofing,'' {\em arXiv preprint arXiv:2011.08019}, 2020.

\bibitem{grosz2020fingerprint}
S.~A. Grosz, T.~Chugh, and A.~K. Jain, ``Fingerprint presentation attack
  detection: A sensor and material agnostic approach,'' in {\em IEEE IJCB},
  pp.~1--10, 2020.

\bibitem{mirzaalian2019effectiveness}
H.~Mirzaalian, M.~Hussein, and W.~Abd-Almageed, ``On the effectiveness of laser
  speckle contrast imaging and deep neural networks for detecting known and
  unknown fingerprint presentation attacks,'' in {\em IEEE ICB}, pp.~1--8,
  2019.

\bibitem{rattani2015open}
A.~Rattani, W.~J. Scheirer, and A.~Ross, ``Open set fingerprint spoof detection
  across novel fabrication materials,'' {\em IEEE TIFS}, vol.~10, no.~11,
  pp.~2447--2460, 2015.

\bibitem{ding2016ensemble}
Y.~Ding and A.~Ross, ``An ensemble of one-class svms for fingerprint spoof
  detection across different fabrication materials,'' in {\em IEEE
  International Workshop on Information Forensics and Security}, pp.~1--6,
  2016.

\bibitem{arashloo2017anomaly}
S.~R. Arashloo, J.~Kittler, and W.~Christmas, ``An anomaly detection approach
  to face spoofing detection: A new formulation and evaluation protocol,'' {\em
  IEEE access}, vol.~5, pp.~13868--13882, 2017.

\bibitem{nikisins2018effectiveness}
O.~Nikisins, A.~Mohammadi, A.~Anjos, and S.~Marcel, ``On effectiveness of
  anomaly detection approaches against unseen presentation attacks in face
  anti-spoofing,'' in {\em IEEE ICB}, pp.~75--81, 2018.

\bibitem{yang2019face}
X.~Yang, W.~Luo, L.~Bao, Y.~Gao, D.~Gong, S.~Zheng, Z.~Li, and W.~Liu, ``Face
  anti-spoofing: Model matters, so does data,'' in {\em CVPR}, pp.~3507--3516,
  2019.

\bibitem{patel2016cross}
K.~Patel, H.~Han, and A.~K. Jain, ``Cross-database face antispoofing with
  robust feature representation,'' in {\em Chinese Conference on Biometric
  Recognition}, pp.~611--619, 2016.

\bibitem{tu2019deep}
X.~Tu, H.~Zhang, M.~Xie, Y.~Luo, Y.~Zhang, and Z.~Ma, ``Deep transfer across
  domains for face antispoofing,'' {\em Journal of Electronic Imaging},
  vol.~28, no.~4, p.~043001, 2019.

\bibitem{jia2020single}
Y.~Jia, J.~Zhang, S.~Shan, and X.~Chen, ``Single-side domain generalization for
  face anti-spoofing,'' in {\em CVPR}, pp.~8484--8493, 2020.

\bibitem{li2018unsupervised}
H.~Li, W.~Li, H.~Cao, S.~Wang, F.~Huang, and A.~C. Kot, ``Unsupervised domain
  adaptation for face anti-spoofing,'' {\em IEEE TIFS}, vol.~13, no.~7,
  pp.~1794--1809, 2018.

\bibitem{wang2019improving}
G.~Wang, H.~Han, S.~Shan, and X.~Chen, ``Improving cross-database face
  presentation attack detection via adversarial domain adaptation,'' in {\em
  IEEE ICB}, pp.~1--8, 2019.

\bibitem{chugh2018fingerprint}
T.~Chugh, K.~Cao, and A.~K. Jain, ``Fingerprint spoof buster: Use of
  minutiae-centered patches,'' {\em IEEE TIFS}, vol.~13, no.~9, pp.~2190--2202,
  2018.

\bibitem{tan2010effect}
B.~Tan, A.~Lewicke, D.~Yambay, and S.~Schuckers, ``The effect of environmental
  conditions and novel spoofing methods on fingerprint anti-spoofing
  algorithms,'' in {\em IEEE International Workshop on Information Forensics
  and Security}, pp.~1--6, 2010.

\bibitem{popli2021unified}
A.~Popli, S.~Tandon, J.~J. Engelsma, N.~Onoe, A.~Okubo, and A.~Namboodiri, ``A
  unified model for fingerprint authentication and presentation attack
  detection,'' {\em arXiv preprint arXiv:2104.03255}, 2021.

\bibitem{faceguard}
D.~Deb, X.~Liu, and A.~K. Jain, ``Faceguard: A self-supervised defense against
  adversarial face images,'' {\em arXiv preprint arXiv:2011.14218}, 2020.

\bibitem{fgsm}
I.~J. Goodfellow, J.~Shlens, and C.~Szegedy, ``Explaining and harnessing
  adversarial examples,'' {\em arXiv preprint arXiv:1412.6572}, 2014.

\bibitem{advfaces}
D.~Deb, J.~Zhang, and A.~K. Jain, ``Advfaces: Adversarial face synthesis,'' in
  {\em IEEE IJCB}, 2020.

\bibitem{dong}
Y.~Dong, H.~Su, B.~Wu, Z.~Li, W.~Liu, T.~Zhang, and J.~Zhu, ``Efficient
  decision-based black-box adversarial attacks on face recognition,'' in {\em
  CVPR}, pp.~7714--7722, 2019.

\bibitem{gflm}
A.~Dabouei, S.~Soleymani, J.~Dawson, and N.~Nasrabadi, ``Fast
  geometrically-perturbed adversarial faces,'' in {\em WACV}, 2019.

\bibitem{semantic_adv}
H.~Qiu, C.~Xiao, L.~Yang, X.~Yan, H.~Lee, and B.~Li, ``Semanticadv: Generating
  adversarial examples via attribute-conditional image editing,'' {\em arXiv
  preprint arXiv:1906.07927}, 2019.

\bibitem{fawkes}
S.~Shan, E.~Wenger, J.~Zhang, H.~Li, H.~Zheng, and B.~Y. Zhao, ``Fawkes:
  Protecting privacy against unauthorized deep learning models,'' in {\em
  USENIX Security Symposium}, 2020.

\bibitem{pgd}
A.~Madry, A.~Makelov, L.~Schmidt, D.~Tsipras, and A.~Vladu, ``Towards deep
  learning models resistant to adversarial attacks,'' {\em arXiv preprint
  arXiv:1706.06083}, 2017.

\bibitem{adv_train}
A.~Kurakin, I.~Goodfellow, and S.~Bengio, ``Adversarial machine learning at
  scale.,'' {\em ICLR}, 2017.

\bibitem{l2l}
Y.~Jang, T.~Zhao, S.~Hong, and H.~Lee, ``Adversarial defense via learning to
  generate diverse attacks,'' in {\em ICCV}, 2019.

\bibitem{feat_denoising}
C.~Xie, Y.~Wu, L.~v.~d. Maaten, A.~L. Yuille, and K.~He, ``Feature denoising
  for improving adversarial robustness,'' in {\em CVPR}, 2019.

\bibitem{robustness_cost1}
D.~Su, H.~Zhang, H.~Chen, J.~Yi, P.-Y. Chen, and Y.~Gao, ``Is robustness the
  cost of accuracy?--a comprehensive study on the robustness of 18 deep image
  classification models.,'' in {\em ECCV}, 2018.

\bibitem{robustness_cost2}
D.~Tsipras, S.~Santurkar, L.~Engstrom, A.~Turner, and A.~Madry, ``Robustness
  may be at odds with accuracy.,'' {\em ICLR}, 2017.

\bibitem{stochastic}
G.~S. Dhillon, K.~Azizzadenesheli, Z.~C. Lipton, J.~Bernstein, J.~Kossaifi,
  A.~Khanna, and A.~Anandkumar, ``Stochastic activation pruning for robust
  adversarial defense,'' in {\em ICLR}, 2018.

\bibitem{artifacts}
R.~Feinman, R.~R. Curtin, S.~Shintre, and A.~B. Gardner, ``Detecting
  adversarial samples from artifacts,'' {\em arXiv preprint arXiv:1703.00410},
  2017.

\bibitem{gong}
Z.~Gong, W.~Wang, and W.-S. Ku, ``Adversarial and clean data are not twins,''
  {\em arXiv preprint arXiv:1704.04960}, 2017.

\bibitem{grosse}
K.~Grosse, P.~Manoharan, N.~Papernot, M.~Backes, and P.~McDaniel, ``On the
  (statistical) detection of adversarial examples,'' {\em arXiv preprint
  arXiv:1702.06280}, 2017.

\bibitem{li_defense}
X.~Li and F.~Li, ``Adversarial examples detection in deep networks with
  convolutional filter statistics,'' in {\em ICCV}, pp.~5764--5772, 2017.

\bibitem{hendrycks}
D.~Hendrycks and K.~Gimpel, ``Early methods for detecting adversarial images,''
  {\em arXiv preprint arXiv:1608.00530}, 2016.

\bibitem{guo}
C.~Guo, M.~Rana, M.~Cisse, and L.~Van Der~Maaten, ``Countering adversarial
  images using input transformations,'' {\em arXiv preprint arXiv:1711.00117},
  2017.

\bibitem{logit_pairing}
H.~Kannan, A.~Kurakin, and I.~Goodfellow, ``Adversarial logit pairing,'' {\em
  arXiv preprint arXiv:1803.06373}, 2018.

\bibitem{metzen}
J.~H. Metzen, T.~Genewein, V.~Fischer, and B.~Bischoff, ``On detecting
  adversarial perturbations,'' {\em ICLR}, 2017.

\bibitem{cascade}
T.~Na, J.~H. Ko, and S.~Mukhopadhyay, ``Cascade adversarial machine learning
  regularized with a unified embedding,'' {\em ICLR}, 2017.

\bibitem{xie}
C.~Xie, J.~Wang, Z.~Zhang, Z.~Ren, and A.~Yuille, ``Mitigating adversarial
  effects through randomization,'' {\em ICLR}, 2017.

\bibitem{efficient}
V.~Zantedeschi, M.-I. Nicolae, and A.~Rawat, ``Efficient defenses against
  adversarial attacks,'' in {\em ACM Workshop on Artificial Intelligence and
  Security}, pp.~39--49, 2017.

\bibitem{uapd}
A.~Agarwal, R.~Singh, M.~Vatsa, and N.~Ratha, ``Are image-agnostic universal
  adversarial perturbations for face recognition difficult to detect?,'' in
  {\em BTAS}, pp.~1--7, 2018.

\bibitem{smartbox}
A.~Goel, A.~Singh, A.~Agarwal, M.~Vatsa, and R.~Singh, ``Smartbox: Benchmarking
  adversarial detection and mitigation algorithms for face recognition.,'' in
  {\em BTAS}, pp.~1--7, 2018.

\bibitem{massoli}
F.~V. Massoli, F.~Carrara, G.~Amato, and F.~Falchi, ``Detection of face
  recognition adversarial attacks,'' {\em CVIU}, p.~103103, 2020.

\bibitem{massoli-cross}
F.~V. Massoli, F.~Falchi, and G.~Amato, ``Cross-resolution face recognition
  adversarial attacks,'' {\em Pattern Recognition Letters}, vol.~140,
  pp.~222--229, 2020.

\bibitem{agarwal_image_transform}
A.~Agarwal, R.~Singh, M.~Vatsa, and N.~K. Ratha, ``Image transformation based
  defense against adversarial perturbation on deep learning models,'' {\em IEEE
  Dependable and Secure Computing}, 2020.

\bibitem{goswami2019detecting}
G.~Goswami, A.~Agarwal, N.~Ratha, R.~Singh, and M.~Vatsa, ``Detecting and
  mitigating adversarial perturbations for robust face recognition,'' {\em
  International Journal of Computer Vision}, vol.~127, no.~6-7, pp.~719--742,
  2019.

\bibitem{singh2020robustness}
R.~Singh, A.~Agarwal, M.~Singh, S.~Nagpal, and M.~Vatsa, ``On the robustness of
  face recognition algorithms against attacks and bias,'' {\em arXiv preprint
  arXiv:2002.02942}, 2020.

\bibitem{magnet}
D.~Meng and H.~Chen, ``Magnet: a two-pronged defense against adversarial
  examples,'' in {\em ACM Conference on Computer and Communications Security},
  pp.~135--147, 2017.

\bibitem{defense_gan}
P.~Samangouei, M.~Kabkab, and R.~Chellappa, ``Defense-gan: Protecting
  classifiers against adversarial attacks using generative models,'' {\em
  ICLR}, 2018.

\bibitem{pixeldefend}
Y.~Song, T.~Kim, S.~Nowozin, S.~Ermon, and N.~Kushman, ``Pixeldefend:
  Leveraging generative models to understand and defend against adversarial
  examples,'' {\em ICLR}, 2017.

\bibitem{self_supervised}
M.~Naseer, S.~Khan, M.~Hayat, F.~S. Khan, and F.~Porikli, ``A self-supervised
  approach for adversarial robustness,'' in {\em CVPR}, 2020.

\bibitem{feat_distillation}
Z.~Liu, Q.~Liu, T.~Liu, N.~Xu, X.~Lin, Y.~Wang, and W.~Wen, ``Feature
  distillation: Dnn-oriented jpeg compression against adversarial examples,''
  in {\em CVPR}, 2019.

\bibitem{avae}
J.~Zhou, C.~Liang, and J.~Chen, ``Manifold projection for adversarial defense
  on face recognition,'' in {\em ECCV}, pp.~288--305, 2020.

\bibitem{mai2018reconstruction}
G.~Mai, K.~Cao, P.~C. Yuen, and A.~K. Jain, ``On the reconstruction of face
  images from deep face templates,'' {\em IEEE PAMI}, vol.~41, no.~5,
  pp.~1188--1202, 2018.

\bibitem{cao2014learning}
K.~Cao and A.~K. Jain, ``Learning fingerprint reconstruction: From minutiae to
  image,'' {\em IEEE TIFS}, vol.~10, no.~1, pp.~104--117, 2014.

\bibitem{ahmad2020resist}
S.~Ahmad and B.~Fuller, ``Resist: Reconstruction of irises from templates,'' in
  {\em IEEE IJCB}, pp.~1--10, 2020.

\bibitem{li2019learning}
R.~Li, D.~Song, Y.~Liu, and J.~Feng, ``Learning global fingerprint features by
  training a fully convolutional network with local patches,'' in {\em IEEE
  ICB}, pp.~1--8, 2019.

\bibitem{nguyen2017iris}
K.~Nguyen, C.~Fookes, A.~Ross, and S.~Sridharan, ``Iris recognition with
  off-the-shelf cnn features: A deep learning perspective,'' {\em IEEE Access},
  vol.~6, pp.~18848--18855, 2017.

\bibitem{tang2017fingernet}
Y.~Tang, F.~Gao, J.~Feng, and Y.~Liu, ``Fingernet: An unified deep network for
  fingerprint minutiae extraction,'' in {\em IEEE IJCB}, pp.~108--116, 2017.

\bibitem{fei2020adversarial}
J.~Fei, Z.~Xia, P.~Yu, and F.~Xiao, ``Adversarial attacks on fingerprint
  liveness detection,'' {\em EURASIP Journal on Image and Video Processing},
  vol.~2020, no.~1, pp.~1--11, 2020.

\bibitem{marrone2021fingerprint}
S.~Marrone, R.~Casula, G.~Orr{\`u}, G.~L. Marcialis, and C.~Sansone,
  ``Fingerprint adversarial presentation attack in the physical domain,'' in
  {\em IEEE ICPR}, pp.~530--543, 2021.

\bibitem{jassim2009improving}
S.~Jassim, H.~Al-Assam, and H.~Sellahewa, ``Improving performance and security
  of biometrics using efficient and stable random projection techniques,'' in
  {\em IEEE International Symposium on Image and Signal Processing and
  Analysis}, pp.~556--561, 2009.

\bibitem{soleymani2019adversarial}
S.~Soleymani, A.~Dabouei, J.~Dawson, and N.~M. Nasrabadi, ``Adversarial
  examples to fool iris recognition systems,'' in {\em IEEE ICB}, pp.~1--8,
  2019.

\bibitem{soleymani2019defending}
S.~Soleymani, A.~Dabouei, J.~Dawson, and N.~M. Nasrabadi, ``Defending against
  adversarial iris examples using wavelet decomposition,'' in {\em IEEE BTAS},
  pp.~1--9, 2019.

\bibitem{tamizhiniyan2021deepiris}
S.~Tamizhiniyan, A.~Ojha, K.~Meenakshi, and G.~Maragatham, ``Deepiris: An
  ensemble approach to defending iris recognition classifiers against
  adversarial attacks,'' in {\em IEEE International Conference on Computer
  Communication and Informatics}, pp.~1--8, 2021.

\bibitem{feng2010fingerprint}
J.~Feng and A.~K. Jain, ``Fingerprint reconstruction: from minutiae to phase,''
  {\em IEEE PAMI}, vol.~33, no.~2, pp.~209--223, 2010.

\bibitem{galbally2012iriscode}
J.~Galbally, A.~Ross, M.~Gomez-Barrero, J.~Fierrez, and J.~Ortega-Garcia,
  ``From the iriscode to the iris: A new vulnerability of iris recognition
  systems,'' {\em Black Hat Briefings USA}, vol.~1, 2012.

\bibitem{galbally2013iris}
J.~Galbally, A.~Ross, M.~Gomez-Barrero, J.~Fierrez, and J.~Ortega-Garcia,
  ``Iris image reconstruction from binary templates: An efficient probabilistic
  approach based on genetic algorithms,'' {\em Computer Vision and Image
  Understanding}, vol.~117, no.~10, pp.~1512--1525, 2013.

\bibitem{dhar2020attributes}
P.~Dhar, A.~Bansal, C.~D. Castillo, J.~Gleason, P.~J. Phillips, and
  R.~Chellappa, ``How are attributes expressed in face dcnns?,'' in {\em IEEE
  FG}, pp.~85--92, 2020.

\bibitem{terhorst2021comprehensive}
P.~Terh{\"o}rst, J.~N. Kolf, M.~Huber, F.~Kirchbuchner, N.~Damer, A.~Morales,
  J.~Fierrez, and A.~Kuijper, ``A comprehensive study on face recognition
  biases beyond demographics,'' {\em arXiv preprint arXiv:2103.01592}, 2021.

\bibitem{upmanyu2010blind}
M.~Upmanyu, A.~M. Namboodiri, K.~Srinathan, and C.~Jawahar, ``Blind
  authentication: a secure crypto-biometric verification protocol,'' {\em IEEE
  TIFS}, vol.~5, no.~2, pp.~255--268, 2010.

\bibitem{fuzzy1}
U.~Uludag, S.~Pankanti, and A.~K. Jain, ``Fuzzy vault for fingerprints,'' in
  {\em International Conference on Audio-and Video-Based Biometric Person
  Authentication}, pp.~310--319, 2005.

\bibitem{fuzzy2}
Y.~J. Lee, K.~R. Park, S.~J. Lee, K.~Bae, and J.~Kim, ``A new method for
  generating an invariant iris private key based on the fuzzy vault system,''
  {\em IEEE Systems, Man, and Cybernetics}, vol.~38, no.~5, pp.~1302--1313,
  2008.

\bibitem{ratha2001enhancing}
N.~K. Ratha, J.~H. Connell, and R.~M. Bolle, ``Enhancing security and privacy
  in biometrics-based authentication systems,'' {\em IBM systems Journal},
  vol.~40, no.~3, pp.~614--634, 2001.

\bibitem{patel2015cancelable}
V.~M. Patel, N.~K. Ratha, and R.~Chellappa, ``Cancelable biometrics: A
  review,'' {\em IEEE Signal Processing Magazine}, vol.~32, no.~5, pp.~54--65,
  2015.

\bibitem{pr1}
K.~Nandakumar, A.~Nagar, and A.~K. Jain, ``Hardening fingerprint fuzzy vault
  using password,'' in {\em IEEE ICB}, 2007.

\bibitem{pr2}
V.~N. Boddeti and B.~V. Kumar, ``A framework for binding and retrieving
  class-specific information to and from image patterns using correlation
  filters,'' {\em IEEE PAMI}, vol.~35, no.~9, pp.~2064--2077, 2012.

\bibitem{mai2020secureface}
G.~Mai, K.~Cao, X.~Lan, and P.~C. Yuen, ``Secureface: Face template
  protection,'' {\em IEEE TIFS}, vol.~16, pp.~262--277, 2020.

\bibitem{kim2021ironmask}
S.~Kim, Y.~Jeong, J.~Kim, J.~Kim, H.~T. Lee, and J.~H. Seo, ``Ironmask: Modular
  architecture for protecting deep face template,'' in {\em CVPR}, 2021.

\bibitem{he1}
M.~Barni, G.~Droandi, and R.~Lazzeretti, ``Privacy protection in
  biometric-based recognition systems: A marriage between cryptography and
  signal processing,'' {\em IEEE Signal Processing Magazine}, vol.~32, no.~5,
  pp.~66--76, 2015.

\bibitem{he2}
R.~L. Lagendijk, Z.~Erkin, and M.~Barni, ``Encrypted signal processing for
  privacy protection: Conveying the utility of homomorphic encryption and
  multiparty computation,'' {\em IEEE Signal Processing Magazine}, vol.~30,
  no.~1, pp.~82--105, 2012.

\bibitem{he3}
M.~Gomez-Barrero, E.~Maiorana, J.~Galbally, P.~Campisi, and J.~Fierrez,
  ``Multi-biometric template protection based on homomorphic encryption,'' {\em
  Pattern Recognition}, vol.~67, pp.~149--163, 2017.

\bibitem{he4}
J.~H. Cheon, H.~Chung, M.~Kim, and K.-W. Lee, ``Ghostshell: Secure biometric
  authentication using integrity-based homomorphic evaluations.,'' {\em IACR
  Cryptology ePrint Archive}, vol.~2016, p.~484, 2016.

\bibitem{he5}
J.~Kolberg, P.~Drozdowski, M.~Gomez-Barrero, C.~Rathgeb, and C.~Busch,
  ``Efficiency analysis of post-quantum-secure face template protection schemes
  based on homomorphic encryption,'' in {\em IEEE BIOSIG}, pp.~1--4, 2020.

\bibitem{fhe1}
J.~R. Troncoso-Pastoriza, D.~Gonz{\'a}lez-Jim{\'e}nez, and
  F.~P{\'e}rez-Gonz{\'a}lez, ``Fully private noninteractive face
  verification,'' {\em IEEE TIFS}, vol.~8, no.~7, pp.~1101--1114, 2013.

\bibitem{fhe2}
V.~N. Boddeti, ``Secure face matching using fully homomorphic encryption,'' in
  {\em IEEE BTAS}, 2018.

\bibitem{fhe3}
J.~J. Engelsma, A.~K. Jain, and V.~N. Boddeti, ``Hers: Homomorphically
  encrypted representation search,'' {\em arXiv preprint arXiv:2003.12197},
  2020.

\bibitem{unifad}
D.~Deb, X.~Liu, and A.~K. Jain, ``Unified detection of digital and physical
  face attacks,'' {\em arXiv preprint arXiv:2104.02156}, 2021.

\bibitem{yin2019towards}
B.~Yin, L.~Tran, H.~Li, X.~Shen, and X.~Liu, ``Towards interpretable face
  recognition,'' in {\em ICCV}, pp.~9348--9357, 2019.

\bibitem{innocent2019}
``Ten years later: The lasting impact of the 2009 nas report, 2019..''
  \url{https://www.innocenceproject.org/lasting-impact-of-2009-nas-report/}.

\bibitem{committee2009strengthening}
C.~on~Identifying the Needs of~the Forensic Sciences~Community, N.~R. C. U.~C.
  on~Science, L.~Policy, G.~Affairs, C.~on~Science, Law, C.~on~Applied, and
  T.~Statistics, {\em Strengthening forensic science in the United States: a
  path forward}.
\newblock National Academy Press, 2009.

\bibitem{zeiler2014visualizing}
M.~D. Zeiler and R.~Fergus, ``Visualizing and understanding convolutional
  networks,'' in {\em ECCV}, pp.~818--833, 2014.

\bibitem{mahendran2015understanding}
A.~Mahendran and A.~Vedaldi, ``Understanding deep image representations by
  inverting them,'' in {\em CVPR}, 2015.

\bibitem{yosinski2015understanding}
J.~Yosinski, J.~Clune, A.~Nguyen, T.~Fuchs, and H.~Lipson, ``Understanding
  neural networks through deep visualization,'' {\em arXiv preprint
  arXiv:1506.06579}, 2015.

\bibitem{dosovitskiy2016inverting}
A.~Dosovitskiy and T.~Brox, ``Inverting visual representations with
  convolutional networks,'' in {\em CVPR}, pp.~4829--4837, 2016.

\bibitem{olah2017feature}
C.~Olah, A.~Mordvintsev, and L.~Schubert, ``Feature visualization,'' {\em
  Distill}, vol.~2, no.~11, p.~e7, 2017.

\bibitem{sundararajan2017axiomatic}
M.~Sundararajan, A.~Taly, and Q.~Yan, ``Axiomatic attribution for deep
  networks,'' in {\em ICML}, pp.~3319--3328, 2017.

\bibitem{lundberg2017unified}
S.~Lundberg and S.-I. Lee, ``A unified approach to interpreting model
  predictions,'' {\em arXiv preprint arXiv:1705.07874}, 2017.

\bibitem{shrikumar2017learning}
A.~Shrikumar, P.~Greenside, and A.~Kundaje, ``Learning important features
  through propagating activation differences,'' in {\em ICML}, pp.~3145--3153,
  2017.

\bibitem{simonyan2013deep}
K.~Simonyan, A.~Vedaldi, and A.~Zisserman, ``Deep inside convolutional
  networks: Visualising image classification models and saliency maps,'' {\em
  arXiv preprint arXiv:1312.6034}, 2013.

\bibitem{fong2017interpretable}
R.~C. Fong and A.~Vedaldi, ``Interpretable explanations of black boxes by
  meaningful perturbation,'' in {\em CVPR}, pp.~3429--3437, 2017.

\bibitem{zhou2016learning}
B.~Zhou, A.~Khosla, A.~Lapedriza, A.~Oliva, and A.~Torralba, ``Learning deep
  features for discriminative localization,'' in {\em CVPR}, pp.~2921--2929,
  2016.

\bibitem{selvaraju2017grad}
R.~R. Selvaraju, M.~Cogswell, A.~Das, R.~Vedantam, D.~Parikh, and D.~Batra,
  ``Grad-cam: Visual explanations from deep networks via gradient-based
  localization,'' in {\em ICCV}, pp.~618--626, 2017.

\bibitem{smilkov2017smoothgrad}
D.~Smilkov, N.~Thorat, B.~Kim, F.~Vi{\'e}gas, and M.~Wattenberg, ``Smoothgrad:
  removing noise by adding noise,'' {\em arXiv preprint arXiv:1706.03825},
  2017.

\bibitem{stylianou2019visualizing}
A.~Stylianou, R.~Souvenir, and R.~Pless, ``Visualizing deep similarity
  networks,'' in {\em WACV}, pp.~2029--2037, 2019.

\bibitem{chenexplainable}
C.~Chen and A.~Ross, ``An explainable attention-guided iris presentation attack
  detector,'' in {\em WACV}, 2021.

\bibitem{shi2019probabilistic}
Y.~Shi and A.~K. Jain, ``Probabilistic face embeddings,'' in {\em ICCV}, 2019.

\bibitem{chowdhury2020can}
A.~Chowdhury, S.~Kirchgasser, A.~Uhl, and A.~Ross, ``Can a cnn automatically
  learn the significance of minutiae points for fingerprint matching?,'' in
  {\em WACV}, pp.~351--359, 2020.

\bibitem{liu2020disentangling}
Y.~Liu, J.~Stehouwer, and X.~Liu, ``On disentangling spoof trace for generic
  face anti-spoofing,'' in {\em ECCV}, pp.~406--422, 2020.

\bibitem{liu2020physics}
Y.~Liu and X.~Liu, ``Physics-guided spoof trace disentanglement for generic
  face anti-spoofing,'' {\em arXiv preprint arXiv:2012.05185}, 2020.

\bibitem{sharma2020d}
R.~Sharma and A.~Ross, ``D-netpad: An explainable and interpretable iris
  presentation attack detector,'' in {\em IEEE IJCB}, 2020.

\bibitem{nyt_bias}
{New York Times}, ``{Wrongfully Accused by an Algorithm}.''
  \url{https://www.nytimes.com/2020/06/24/technology/facial-recognition-arrest.html},
  2020.
\newblock [Online; accessed 9-April-2021].

\bibitem{howard2019effect}
J.~Howard, Y.~Sirotin, and A.~Vemury, ``The effect of broad and specific
  demographic homogeneity on the imposter distributions and false match rates
  in face recognition algorithm performance,'' in {\em IEEE BTAS}, 2019.

\bibitem{gender_shades}
M.~M. Lab, ``Algorithmic bias persists.''
  \url{https://www.media.mit.edu/projects/gender-shades/overview}, 2021.
\newblock [Online; accessed 7-May-2021].

\bibitem{news-mit}
https://news.mit.edu/2018/study-finds-gender-skin-type-bias-artificial-intelligence-systems-0212.

\bibitem{gong2020jointly}
S.~Gong, X.~Liu, and A.~K. Jain, ``Jointly de-biasing face recognition and
  demographic attribute estimation,'' {\em ECCV}, 2020.

\bibitem{jain2016fingerprint}
A.~K. Jain, S.~S. Arora, K.~Cao, L.~Best-Rowden, and A.~Bhatnagar,
  ``Fingerprint recognition of young children,'' {\em IEEE TIFS}, vol.~12,
  no.~7, pp.~1501--1514, 2016.

\bibitem{j2019infant}
J.~J~Engelsma, D.~Deb, A.~Jain, A.~Bhatnagar, and P.~Sewak~Sudhish,
  ``Infant-prints: Fingerprints for reducing infant mortality,'' in {\em CVPR
  Workshop}, pp.~67--74, 2019.

\bibitem{fang2021demographic}
M.~Fang, N.~Damer, F.~Kirchbuchner, and A.~Kuijper, ``Demographic bias in
  presentation attack detection of iris recognition systems,'' in {\em IEEE
  EUSIPCO}, 2021.

\bibitem{buolamwini2018gender}
J.~Buolamwini and T.~Gebru, ``Gender shades: Intersectional accuracy
  disparities in commercial gender classification,'' in {\em Conference on
  fairness, accountability and transparency}, 2018.

\bibitem{raji2019actionable}
I.~D. Raji and J.~Buolamwini, ``Actionable auditing: Investigating the impact
  of publicly naming biased performance results of commercial ai products,'' in
  {\em AAAI}, 2019.

\bibitem{ross2004biometric}
A.~Ross and A.~Jain, ``Biometric sensor interoperability: A case study in
  fingerprints,'' in {\em International Workshop on Biometric Authentication},
  pp.~134--145, Springer, 2004.

\bibitem{cook2019demographic}
C.~M. Cook, J.~J. Howard, Y.~B. Sirotin, J.~L. Tipton, and A.~R. Vemury,
  ``Demographic effects in facial recognition and their dependence on image
  acquisition: An evaluation of eleven commercial systems,'' {\em IEEE
  Transactions on Biometrics, Behavior, and Identity Science}, vol.~1, no.~1,
  pp.~32--41, 2019.

\bibitem{osoba2017intelligence}
O.~A. Osoba and W.~Welser~IV, {\em An intelligence in our image: The risks of
  bias and errors in artificial intelligence}.
\newblock Santa Monica, CA, USA:Rand Corporation, 2017.

\bibitem{washington2018argue}
A.~L. Washington, ``How to argue with an algorithm: Lessons from the
  compas-propublica debate,'' {\em Colorado Technol. Law J}, vol.~17, p.~131,
  2018.

\bibitem{garvie2016perpetual}
C.~Garvie, {\em The perpetual line-up: Unregulated police face recognition in
  America}.
\newblock Georgetown Law, Center on Privacy \& Technology, 2016.

\bibitem{o2016weapons}
C.~O'neil, {\em Weapons of math destruction: How big data increases inequality
  and threatens democracy}.
\newblock Crown, 2016.

\bibitem{wang2019racial}
M.~Wang, W.~Deng, J.~Hu, X.~Tao, and Y.~Huang, ``Racial faces in the wild:
  Reducing racial bias by information maximization adaptation network,'' in
  {\em ICCV}, 2019.

\bibitem{nalini1}
N.~Jain, K.~Nandakumar, N.~Ratha, S.~Pankanti, and U.~Kumar, ``Efficient cnn
  building blocks for encrypted data,'' {\em arXiv preprint arXiv:2102.00319},
  2021.

\bibitem{nalini2}
K.~Sarpatwar, K.~Nandakumar, N.~Ratha, J.~Rayfield, K.~Shanmugam, S.~Pankanti,
  and R.~Vaculin, ``Efficient encrypted inference on ensembles of decision
  trees,'' {\em arXiv preprint arXiv:2103.03411}, 2021.

\bibitem{nalini3}
K.~Nandakumar, N.~Ratha, S.~Pankanti, and S.~Halevi, ``Towards deep neural
  network training on encrypted data,'' in {\em CVPR}, pp.~0--0, 2019.

\bibitem{gilad2016cryptonets}
R.~Gilad-Bachrach, N.~Dowlin, K.~Laine, K.~Lauter, M.~Naehrig, and J.~Wernsing,
  ``{CryptoNets}: Applying neural networks to encrypted data with high
  throughput and accuracy,'' in {\em ICML}, 2016.

\bibitem{brutzkus2019low}
A.~Brutzkus, R.~Gilad-Bachrach, and O.~Elisha, ``Low latency privacy preserving
  inference,'' in {\em ICML}, pp.~812--821, 2019.

\bibitem{yonetani2017privacy}
R.~Yonetani, V.~Naresh~Boddeti, K.~M. Kitani, and Y.~Sato, ``Privacy-preserving
  visual learning using doubly permuted homomorphic encryption,'' in {\em
  ICCV}, 2017.

\bibitem{fedface}
D.~Aggarwal, J.~Zhou, and A.~K. Jain, ``Fedface: Collaborative learning of face
  recognition model,'' {\em arXiv preprint arXiv:2104.03008}, 2021.

\bibitem{drummond2003c4}
C.~Drummond, R.~C. Holte, {\em et~al.}, ``C4. 5, class imbalance, and cost
  sensitivity: why under-sampling beats over-sampling,'' in {\em Workshop on
  Learning from Imbalanced Datasets II}, 2003.

\bibitem{chawla2002smote}
N.~V. Chawla, K.~W. Bowyer, L.~O. Hall, and W.~P. Kegelmeyer, ``Smote:
  synthetic minority over-sampling technique,'' {\em Journal of Artificial
  Intelligence research}, vol.~16, pp.~321--357, 2002.

\bibitem{mullick2019generative}
S.~S. Mullick, S.~Datta, and S.~Das, ``Generative adversarial minority
  oversampling,'' {\em arXiv preprint arXiv:1903.09730}, 2019.

\bibitem{cao2019learning}
K.~Cao, C.~Wei, A.~Gaidon, N.~Arechiga, and T.~Ma, ``Learning imbalanced
  datasets with label-distribution-aware margin loss,'' {\em arXiv preprint
  arXiv:1906.07413}, 2019.

\bibitem{cui2019class}
Y.~Cui, M.~Jia, T.-Y. Lin, Y.~Song, and S.~Belongie, ``Class-balanced loss
  based on effective number of samples,'' in {\em CVPR}, 2019.

\bibitem{wang2020mitigating}
M.~Wang and W.~Deng, ``Mitigating bias in face recognition using skewness-aware
  reinforcement learning,'' in {\em CVPR}, 2020.

\bibitem{gong2021mitigating}
S.~Gong, X.~Liu, and A.~K. Jain, ``Mitigating face recognition bias via group
  adaptive classifier,'' in {\em CVPR}, 2021.

\bibitem{marasco2019biases}
E.~Marasco, ``Biases in fingerprint recognition systems: Where are we at?,'' in
  {\em IEEE BTAS}, 2019.

\bibitem{eu_gdpr}
``The general data protection regulation (eu).'' \url{https://bit.ly/3t3BqjW}.

\bibitem{clearview}
``Clearview ai uses your online photos to instantly id you. that’s a problem,
  lawsuit says.'' \url{https://lat.ms/3uqRn3Z}.

\bibitem{MSceleb_retracted}
``Microsoft quietly deletes largest public face recognition data set.''
  \url{https://on.ft.com/2PA595J}.

\bibitem{duke_mtmc}
``Duke mtmc dataset.'' \url{https://exposing.ai/duke_mtmc/}.

\bibitem{brainwash}
``Brainwash dataset.'' \url{https://exposing.ai/brainwash/}.

\bibitem{watson1992nist}
C.~I. Watson and C.~L. Wilson, ``Nist special database 4,'' {\em NIST},
  vol.~17, no.~77, p.~5, 1992.

\bibitem{watson1993nist}
C.~I. Watson, ``Nist special database 14,'' {\em NIST}, 1993.

\bibitem{tinsley2021face}
P.~Tinsley, A.~Czajka, and P.~Flynn, ``This face does not exist... but it might
  be yours! identity leakage in generative models,'' in {\em WACV},
  pp.~1320--1328, 2021.

\bibitem{shao2020federated}
R.~Shao, P.~Perera, P.~C. Yuen, and V.~M. Patel, ``Federated face presentation
  attack detection,'' {\em arXiv preprint arXiv:2005.14638}, 2020.

\bibitem{ijba}
B.~F. Klare, B.~Klein, E.~Taborsky, A.~Blanton, J.~Cheney, K.~Allen,
  P.~Grother, A.~Mah, and A.~K. Jain, ``Pushing the frontiers of unconstrained
  face detection and recognition: Iarpa janus benchmark a,'' in {\em CVPR},
  pp.~1931--1939, 2015.

\end{thebibliography}
